\documentclass[journal,twoside]{IEEEtran} 
\ifCLASSINFOpdf
   \usepackage[pdftex]{graphicx}
\else
\fi
\usepackage{amsfonts} 
\usepackage{cite} 
\usepackage{xcolor} 
\usepackage{hyperref}
\usepackage{amssymb} 
\usepackage[cmex10]{amsmath}

\usepackage{subfiles}
\begin{document}
%
\title{Rethinking CNN-Based Pansharpening: Guided Colorization of Panchromatic Images via GANs}
%
%
%

\author{Furkan Ozcelik, Ugur Alganci, Elif Sertel, and Gozde Unal
\thanks{F. Ozcelik and G. Unal are with Istanbul Technical University (ITU), Department of Computer Engineering, and ITU-AI: Artificial Intelligence and Data Science Research and Application Center.}
\thanks{U. Alganci and E. Sertel are with Istanbul Technical University (ITU), Department of Geomatics Engineering and ITU - CSCRS (Application and Research  Center for Satellite Communications and Remote Sensing).}
\thanks{\copyright 2020 IEEE.  Personal use of this material is permitted.  Permission from IEEE must be obtained for all other uses, in any current or future media, including reprinting/republishing this material for advertising or promotional purposes, creating new collective works, for resale or redistribution to servers or lists, or reuse of any copyrighted component of this work in other works.}
}
%
%

\markboth{TO APPEAR IN IEEE TRANSACTIONS ON GEOSCIENCE AND REMOTE SENSING, 2020}
{OZCELIK et al.: RETHINKING CNN-BASED PANSHARPENING: GUIDED COLORIZATION OF PANCHROMATIC IMAGES VIA GANS}
%



\maketitle

\begin{abstract}
 Convolutional Neural Networks (CNN)-based approaches have shown promising results in pansharpening of satellite images in recent years. However, they still exhibit limitations in producing high-quality pansharpening outputs. To that end, we propose a new self-supervised learning framework, where we treat pansharpening as a colorization problem, which brings an entirely novel perspective and solution to the problem compared to existing methods that base their solution solely on producing a super-resolution version of the multispectral image. Whereas CNN-based methods provide a reduced resolution panchromatic image as input to their model along with reduced resolution multispectral images, hence learn to increase their resolution together, we instead provide the grayscale transformed multispectral image as input, and train our model to learn the colorization of the grayscale input. We further address the fixed downscale ratio assumption during training, which does not generalize well to the full-resolution scenario. We introduce a noise injection into the training by randomly varying the downsampling ratios. Those two critical changes, along with the addition of adversarial training in the proposed PanColorization Generative Adversarial Networks (PanColorGAN) framework, help overcome the spatial detail loss and blur problems that are observed in CNN-based pansharpening. The proposed approach outperforms the previous CNN-based and traditional methods as demonstrated in our experiments.
\end{abstract}


%
\IEEEpeerreviewmaketitle

\section{Introduction}
\label{sec:Intro}

\IEEEPARstart{D}{esigning} algorithms to obtain images with high-resolution properties both in spatial and spectral domains is an important task in remote sensing. As a single sensor is not sufficient to get dual-domain high-resolution images, many of the satellites such as Pleiades, GeoEye, Quickbird, and Worldview constellations contain both panchromatic and multispectral sensors. Panchromatic sensors focus on spatial resolution while providing images with a single-band, whereas multispectral sensors focus on spectral resolution while providing multi-band images. The fusion of these two modalities with a prescribed algorithm in order to obtain high-resolution images in both domains is known as pansharpening.

Traditional methods of pansharpening algorithms can be separated mainly into two categories: component substitution based methods and multiresolution analysis methods \cite{ComparisonSurvey}. Component Substitution (CS) methods transform and split multispectral images into spatial and spectral components, then try to replace the spatial component with a component obtained from panchromatic images. Many variants of CS methods such as PCA, IHS, GS, Brovey Transform, BDSD, and PRACS appeared in the literature \cite{ComponentSubstitution}. Multiresolution Analysis (MRA) methods mainly obtain spatial information by first applying a filter to panchromatic images, followed by an injection of the obtained information to multispectral images \cite{MultiResolutionAnalysis}. There are many examples of MRA methods such as the high-pass filtering (HPF), MTF based methods like Generalized Laplacian pyramids with modulation transfer function (MTF-GLP), MTF-GLP with high pass modulation (MTF-GLP-HPM), MTF-based algorithms with spatial principal component analysis (SPCA) and wavelet-based methods like a trous wavelet transform (ATWT), undecimated discrete wavelet transform (UDWT), and proportional additive wavelet intensity method (AWLP) \cite{MRA1,MRA2,MRA3,MTF-GLP,MTF-GLP-HPM,AWLP}.

Recent availability of large datasets, increased computing power, advanced architectures and optimization led the way to the adaptation of deep learning techniques to numerous problems in computer vision as well as in remote sensing. Typically, a dedicated convolutional neural network (CNN) model is built in order to learn specific supervised learning tasks such as classification and detection, and lately to learn unsupervised learning tasks, particularly in image generation problems. 
For the latter, generative models such as Convolutional Autoencoders and Generative Adversarial Networks (GANs) \cite{GAN} are applied to self-supervised image synthesis tasks such as Super-Resolution (SR) \cite{SRGAN} and Colorization \cite{ColorGAN}. The self-supervision in SR models is realized by reducing the resolution of the input ($2 \times$ - $4 \times$ times typically) during training and allowing the network model to learn to increase the resolution of the input images. The reconstruction loss between the output of the network and the original image is calculated, which is used in the optimization of the network parameters. Colorization is another popular self-supervised synthesis task encountered in computer vision. This time, the network tries to learn to colorize grayscale images, which are created from their color counterparts in the training phase. As the network tries to reconstruct original color images, the corresponding loss between the output of the network and the original image is utilized in the network optimization process.

\begin{figure*}[t!]
\centering
\includegraphics[width=15cm]{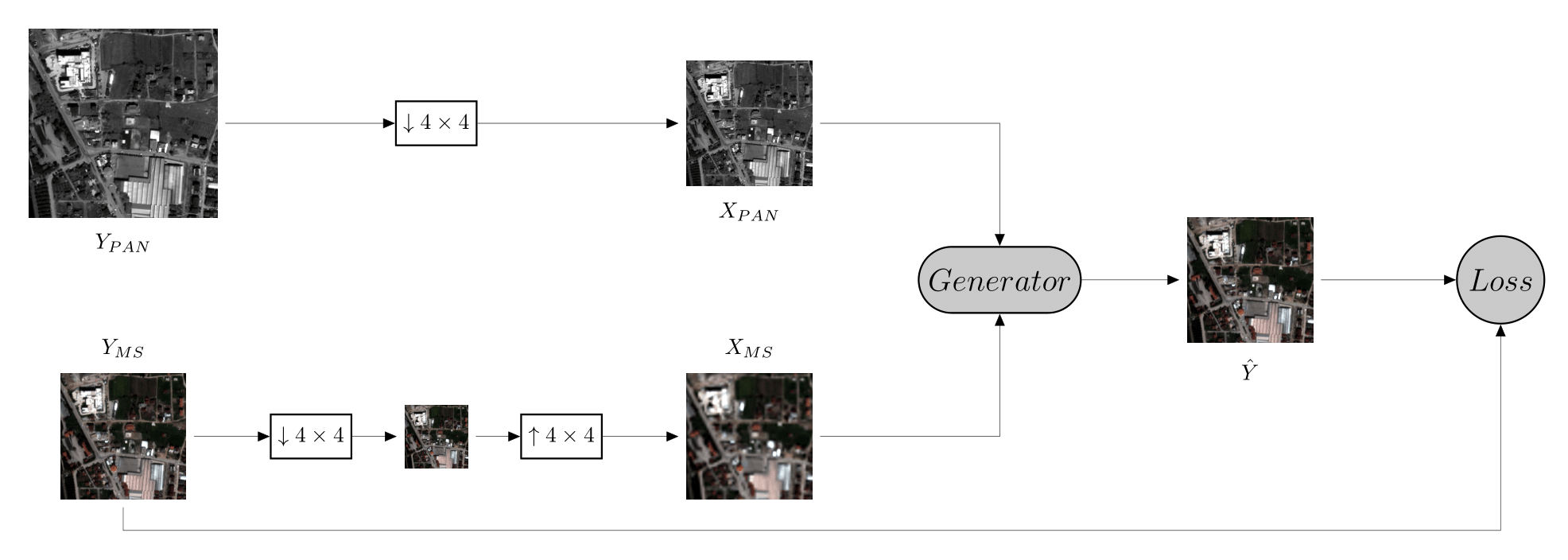}
\caption{Standard CNN-based Pansharpening Framework. Generator refers to a CNN model. A loss function is calculated to train the CNN in order to produce the pansharpened image $\hat{Y}$. }
\label{fig:stdCNN}
\end{figure*}

In the field of remote sensing, in addition to widely-studied supervised learning problems such as land cover classification, building detection, deep network models are recently applied to the pansharpening task \cite{DLRSReview}. Existing CNN-based pansharpening methods in the literature \cite{PNN,PanNet,DNNPan,MRADNN,TFNet,PSGAN,TargetAdaptiveCNN,DResidualPS,BoostingCNN} can be re-interpreted in the framework of self-supervised learning for the super-resolution task while following the commonly used Wald's protocol \cite{WaldProtocol}. Masi et al proposed a three-layer convolutional neural network that is inspired from image super-resolution using deep convolutional neural networks \cite{PNN}. Yang et al proposed a model which uses domain-specific knowledge to enhance structural and spectral properties, while employing high-pass filtering instead of using directly the image \cite{PanNet}. Huang et al used a stacked modified sparse denoising autoencoder for pretraining a deep neural network model effectively \cite{DNNPan}. Liu et al established a model that fuses information gathered from panchromatic and multispectral images at a feature level after several convolution operations \cite{TFNet}. Later, they enhanced the model via a generative adversarial framework by adding a discriminator network \cite{PSGAN}. Scarpa et al utilized a pretrained model that does a fine-tuning on the  target image before the inference stage \cite{TargetAdaptiveCNN}. Wei et al designed a convolutional neural network that uses deep residual learning \cite{BoostingCNN}. In a recent study, Vitale et al devised a cross-scale learning model where it combines losses from both reduced resolution and full resolution comparisons \cite{CS-PNN}. Although they differ in many aspects, all CNN-based methods have some common properties in the training procedure. In the training phase, pansharpening models are provided with the reduced resolution panchromatic and reduced resolution multispectral image as inputs in order to learn to reconstruct a high-resolution multispectral image at the output. Inspired by Wald's protocol, all previous studies treated CNN-based pansharpening only as a super-resolution task. However, we hypothesize and show in this paper that using another self-supervised learning task, namely colorization, is more suitable to the pansharpening problem. 

The motivation behind our introducing a colorization-based self-supervised learning approach to pansharpening is based on our observations of an inefficient level of spatial-detail-preservation in the former approaches. We demonstrate this problem and describe why it is encountered in Section \ref{sec:Motivation}. As a solution, we present a novel pansharpening approach, along with a new GAN-based dedicated colorization model, which we call PanColorGAN in Section \ref{sec:Method}. In Section \ref{sec:Results}, we present the results of the new method, which demonstrates an improved quantitative and qualitative performance, along with discussions, followed by conclusions in Section \ref{sec:Conc}.

\section{Elucidating Why Colorization task is better suited to CNN-based Pansharpening}
\label{sec:Motivation}

In this section, we elucidate issues with the super-resolution based pansharpening approach. First, we describe the standard CNN-based pansharpening framework that is inspired by the super-resolution task in Section \ref{sec:MotivationA}.  In Section \ref{sec:MotivationB}, we present the spatial detail differences across reduced resolution panchromatic images and full resolution multispectral images. We also demonstrate why current pansharpening with deep learning approaches are not efficiently handling this problem in the same section. In Section \ref{subsec:FullResProblems}, we discuss the blurring problem that is caused by an inherent uncertainty in the ratio between full resolution and reduced resolution images.
\vspace*{-0.3cm}
\subsection{Standard CNN-based Pansharpening Framework}
\label{sec:MotivationA}

As stated in Section \ref{sec:Intro}, several pansharpening models were built on CNNs or GANs in the recent literature. Although they offer various architectures, their underlying learning procedures are similar. 
The standard procedure in CNN-based pansharpening methods is based on the Wald's protocol, which is designed to overcome the reference problem in quantitative analysis of pansharpening. In Wald's protocol, the algorithm gets the reduced resolution panchromatic image and the reduced resolution multispectral image as input, and attempts to return an image similar to the original multispectral image as its output through various image processing operations. Deep learning-based models, on the other hand, involve extensive training processes that are designed while adopting Wald's protocol.

\begin{figure}[h]
\centering
\includegraphics[width=9cm]{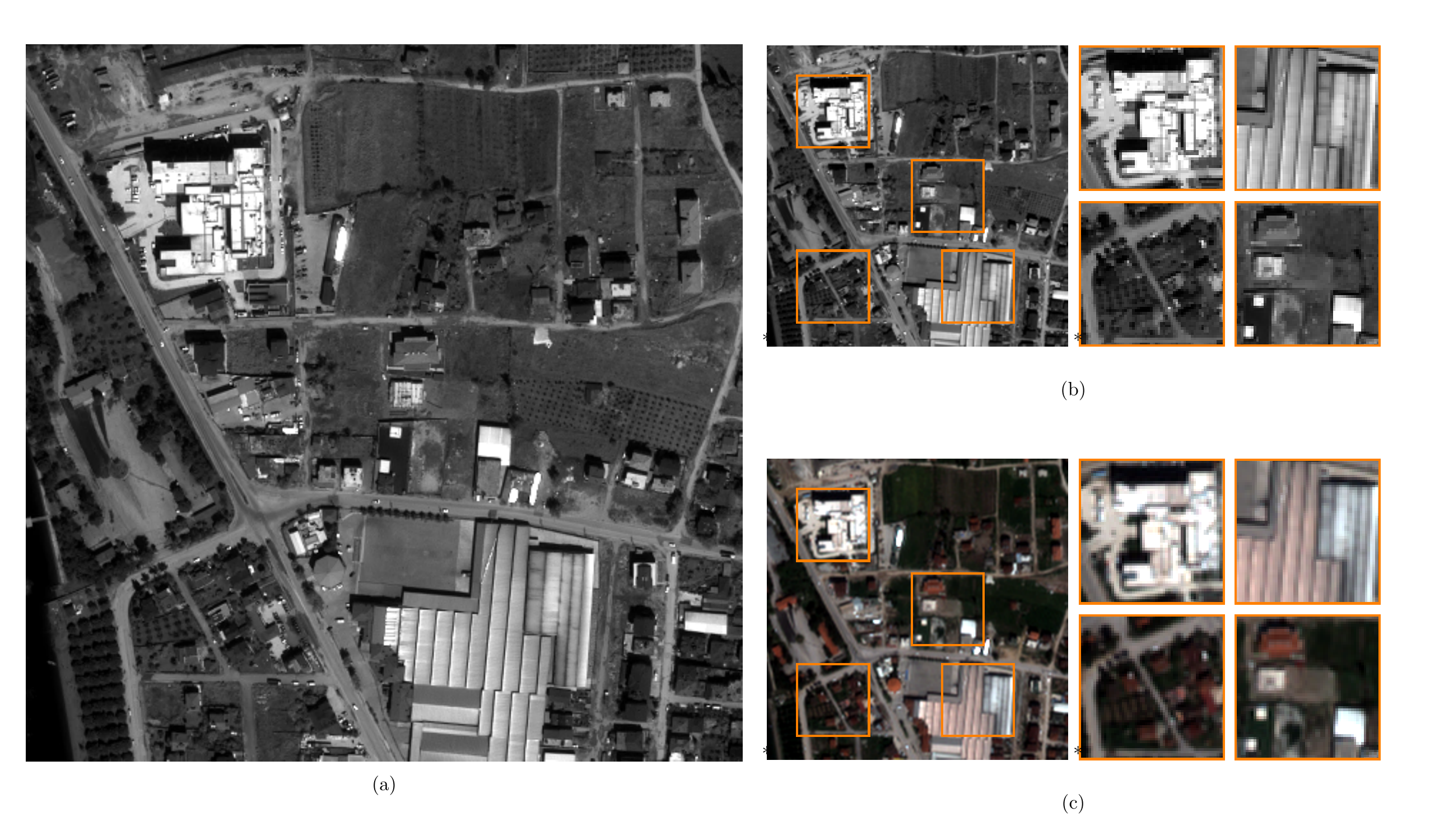} 
\vspace*{-0.6cm}
\caption{Spatial-level-of detail comparison between reduced panchromatic and multispectral images demonstrated on Pleiades dataset. (a) Original panchromatic image. (b) Reduced panchromatic image. (c) Multispectral image. Orange boxes on the left are zoomed into for display on the right. }
\label{fig:sharpness}
\end{figure}

\begin{figure*}[ht]
\centering
\includegraphics[width=17cm]{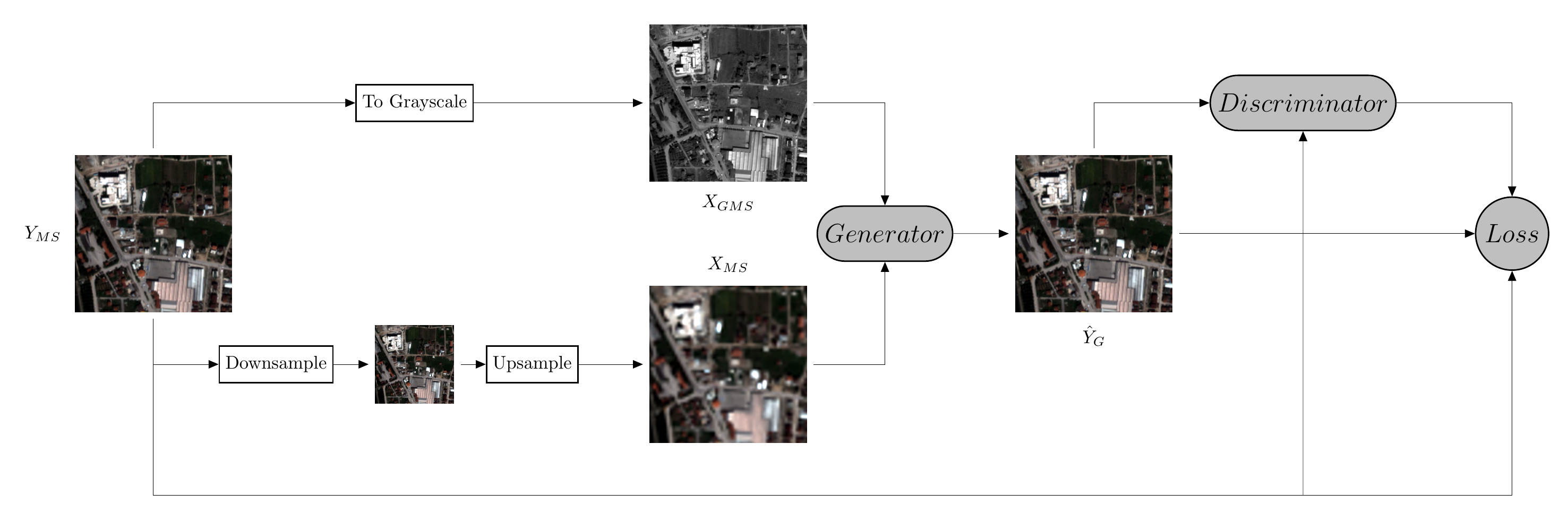}
\caption{Proposed Training Scheme for PanColorGAN Model: A reconstruction loss $Loss(L_1)$ between the colorized output of the $\hat{Y}_G$ input and $Y_{MS}$, as well as an adversarial loss that evaluates the generation quality of $\hat{Y}_G$ generated from $X_{GMS}$ and $X_{MS}$ are utilized to train the PanColorGAN. 
}
\label{fig:PanColorGAN}
\end{figure*}

We illustrate the standard CNN-based pansharpening framework in Figure~\ref{fig:stdCNN}. Suppose that we have $Y_{PAN}$ and $Y_{MS}$, which are corresponding panchromatic (PAN) and multispectral (MS) images that we want to fuse through pansharpening. First, $Y_{PAN}$ is reduced by $4\times$ to the size of the $Y_{MS}$ to obtain the $X_{PAN}$ image. $Y_{MS}$ is reduced by $4\times$, then upsampled by $4\times$ to obtain the $X_{MS}$. $X_{PAN}$ and $X_{MS}$ are provided to a generator network $G$, hence $\hat{Y} = G(X_{PAN},X_{MS})$ is obtained at the output as the generated or pansharpened image. A reconstruction loss function, either with an $L_2$ or $L_1$ norm is calculated between the output and the multispectral image. 

The procedure with standard CNN-based models with or without an adversarial loss then is executed through an optimization of the overall loss function
(see Section~\ref{subsec:GAN}).  Next, we explain the disagreement in spatial details after training such a model.

\subsection{Problems in Reduced Resolution Pansharpening}
\label{sec:MotivationB}
 When one trains a model with the standard CNN-based pansharpening framework, although quantitative results between original multispectral and generated pansharpened images are typically highly favorable, a closer inspection of the inputs and outputs shows that pansharpened images that are obtained from the model do not preserve the desired sharp spatial details that exist in the reduced panchromatic image inputs. We notice that the problem lies within the crucial assumption that the reduced panchromatic images and original multispectral images should have similar spatial quality as they bear the same spatial resolution level. On the contrary, it can be both qualitatively and quantitatively argued that the reduced panchromatic images exhibit better spatial quality than original multispectral images.

\begin{table}[!t]
\renewcommand{\arraystretch}{1.3}
\caption{QUANTITATIVE ANALYSIS OF SPATIAL QUALITY INCOMPATIBILITY BETWEEN REDUCED PANCHROMATIC IMAGES AND MULTISPECTRAL IMAGES}
\label{table_example}
\centering
\begin{tabular}{l c c c}
\hline
 & PSNR & sCC & SSIM \\
 (worst-best) & (0-inf) & (0-1) & (0-1)\\
  \hline
Reduced PAN - Grayscale MS & 24.704 & .088 & .586 \\
Reduced PAN(Blurred) - Grayscale MS & 30.751 & .424 & .848\\

\hline
\end{tabular}
\end{table}

Figure \ref{fig:sharpness} qualitatively demonstrates this problem, where spatial detail disagreement in terms of lack of sharpness in detail, blurriness, reduced contrast differences, and less continuity in lines in the images can be clearly seen by visual inspection (compare zoomed image patches in (b) and (c)). In order to quantitatively test our conjecture, we calculate three measures, which are PSNR, sCC, and SSIM on a set of reduced panchromatic images given the corresponding gray-transformed multispectral images as a reference. Next, we apply a blurring Gaussian filter with  $5\times5$  kernel ($\sigma=2$) to obtain the blurred reduced panchromatic images. We calculate the three measures using this time the blurred panchromatic image rather than the original panchromatic image (Table~\ref{table_example}). Per our hypothesis, quantitative measures should improve with blurred versions of the reduced panchromatic images, since we claim that original multispectral images are blurrier than reduced panchromatic images. It can be observed in Table \ref{table_example} that all three measures change in an expected direction, hence the blurred versions of the reduced panchromatic images show increasingly similar characteristics to multispectral images.

Current deep learning methods used in pansharpening, which are inspired mainly from super-resolution, inherently incorporate the abovementioned spatial detail disagreement issue into their procedures as they involve mapping a function from a pair of reduced resolution panchromatic image and reduced resolution multispectral image to the original multispectral image. Our analysis above shows that reduced resolution panchromatic images contain more spatial details than the original multispectral image, which are lost during the prescribed procedure. This is the main reason behind obtaining decent quantitative results, whereas pansharpened images exhibit reduced spatial details compared to original panchromatic images. 

\subsection{Problems in Full Resolution Pansharpening}
\label{subsec:FullResProblems}
Similar to the reduced resolution procedures, the full-resolution pansharpening procedure is also prone to a specific blurring problem due to the strong assumption of learning a  ``fixed upsample scale'' (e.g. say a typical ratio of $4 \times$) in the training phase of standard CNN-based approaches. As the level of detail of the $4 \times$ reduced resolution panchromatic image may not correspond to that of the original multispectral image, training the CNN-based learning model according to the Wald's Protocol naturally cannot match the desired upsampling ratio exactly, and leads to blurry results for the full resolution case. We present a remedy to that problem, by introducing random downsampling ratios, rather than a fixed (e.g. $4 \times$) reduced scale during training, as the latter does not generalize well to full resolution pansharpening, as is demonstrated in Section~\ref{sec:Results}.

\section{PanSharpening with Guided Colorization using GANs (PanColorGAN)}
\label{sec:Method}
\subsection{Proposed Framework}

\begin{figure*}[ht]
\centering
\includegraphics[width=16cm]{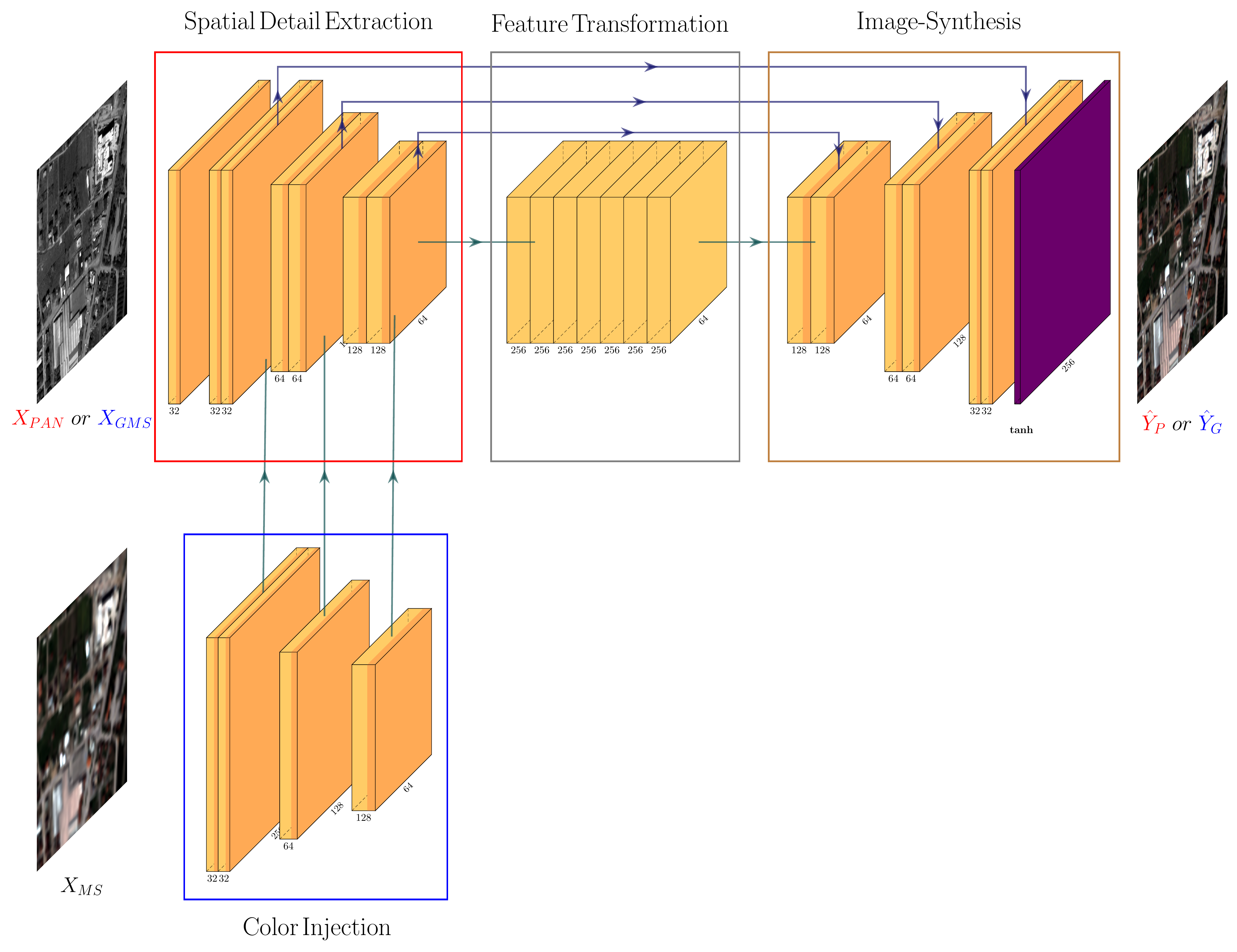}\llap{\includegraphics[width=10cm]{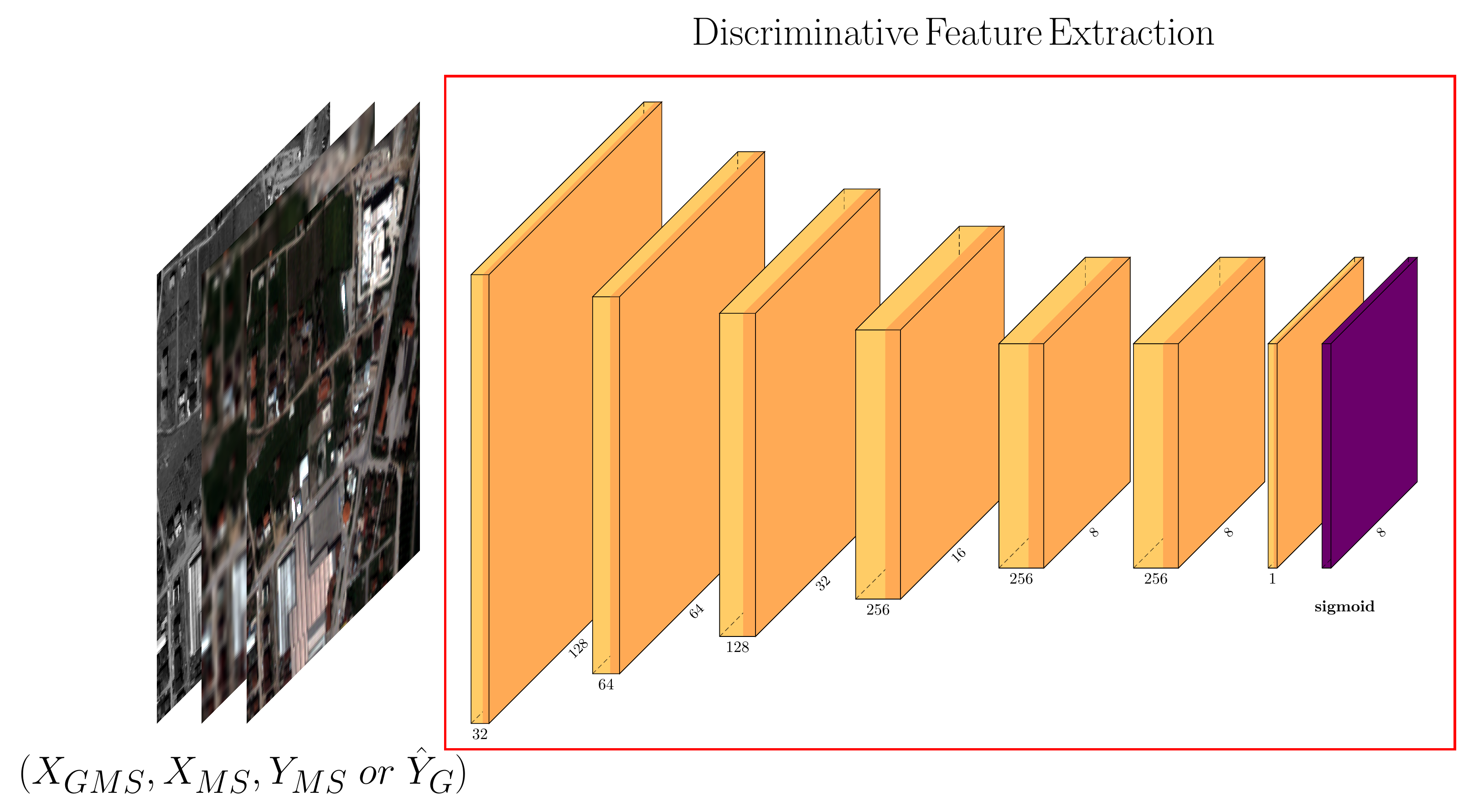}}
\caption{PanColorGAN model: Architecture details for its Generator and Discriminator  Networks are depicted. Two modes exist for Generator network: In the training phase, \textcolor{blue}{$X_{GMS}$} is provided along with $X_{MS}$ to generate \textcolor{blue}{$\hat{Y}_G$}. In the testing phase, \textcolor{red}{$X_{PAN}$} is provided along with $X_{MS}$ to generate \textcolor{red}{$\hat{Y}_P$}. Also during the training phase, Discriminator network gets two different types of batches. A real batch consists a concatenated set of  $X_{GMS}$, $X_{MS}$ and $Y_{MS}$. A fake batch consists a concatenated set of $X_{GMS}$, $X_{MS}$ and $\hat{Y}_G$, as shown on the bottom right. }
\label{fig:PanColorGANArch}
\end{figure*}

To address the shortcomings of the standard CNN-based approaches, we present a new pansharpening method that faithfully preserves spatial details given by the input panchromatic image in the inference stage. This is achieved by designing a self-supervised learning procedure based on the colorization task rather than super-resolution task. This new task that is cast upon the network model requires that during the training phase, we provide the grayscale multispectral image, whose spatial details perfectly agree with those of the original multispectral image. This is not the case for the reduced panchromatic image due to spatial detail disagreement problem that we discussed in Section~\ref{sec:MotivationB}.

To further expound our reasoning on colorization based pansharpening versus super-resolution based pansharpening, an analogy of comparison between traditional CS and MRA methods can be made. Existing super-resolution based pansharpening methods can be considered more similar to MRA methods than CS methods because, in the training phase, the model tries to increase spatial details of reduced resolution multispectral image with spatial features extracted from reduced resolution panchromatic image by comparing it to the  original multispectral image. On the other hand, the colorization based pansharpening method we propose can be interpreted more in line with a CS approach rather than an MRA approach. As we will see more details in the following parts, our model learns to generate an original multispectral image by taking its reduced resolution multispectral image and the corresponding grayscale multispectral image as inputs, which is interpreted as colorization. We can also interpret this in a way that our model learns to separate spectral and spatial components of the multispectral image during training. Then, in the testing stage, we provide the corresponding panchromatic image instead of the grayscale multispectral image, which can be interpreted as substitution of spatial components between two images, which alludes to traditional CS approaches.

Furthermore, we improve the full-resolution pansharpening procedure by injecting noise into the assumed downsampling-upsampling ratios between the original panchromatic and multispectral images, which induces a regularization effect into our model. 

The proposed PanColorGAN pansharpening learning model is illustrated in Figure~\ref{fig:PanColorGAN}. First, let us describe the original PanColorGAN with a fixed down/up-sampling ratio. Suppose that the input multispectral image $Y_{MS}$ is first downsampled by $k=4\times$ then upsampled by $k=4\times$ to obtain $X_{MS}$. $Y_{MS}$ is also transformed to grayscale by taking an average of channels to construct a grayscale input $X_{GMS}$. Later, $X_{GMS}$ and $X_{MS}$ are provided as input to the generator network $G$ and $\hat{Y}_G = G(X_{GMS},X_{MS})$ is obtained as the output. A reconstruction loss is calculated between $\hat{Y}_G$ and $Y_{MS}$. 

In our PanColorGAN, as in traditional GANs, an additional discriminator network $D$ is also built to provide an Adversarial Loss, which is calculated for $\hat{Y}_G$ because we would like to augment the representation capability of the generator network by providing feedback on the quality or the credibility of its generated output. Details of the model are explained next. 

\vspace*{-0.4cm}
\subsection{Adversarial Loss (GAN - RaGAN)}
\label{subsec:GAN}
Generative Adversarial Networks (GANs) belong to the class of generative networks that learn to synthesize images with a target distribution by competition of typically two networks, where one is the generator and the other one is the discriminator \cite{GAN}. In vanilla GANs, the generator $G$ learns to transform a random noise distribution to the target image distribution. Discriminator $D$ aims to correctly classify the output of $G$ with a ``generated'' label versus ``real'' label. Here, the ``real'' refers to a label of the training data. $D$ also performs the same operation on images generated by the model. 
This is the basis of the adversarial loss in the vanilla GAN~\cite{GAN}, which is used in update of both $G$ and $D$:
\begin{eqnarray}
  L_{G A N}^{D}&=&-\mathbb{E}_{x_{r} \sim \mathbb{P}} \log(D(x_r))]-\mathbb{E}_{x_{f} \sim \mathbb{Q}} [\log  (1-D(x_f) ) ]   \nonumber \\
  L_{G A N}^{G}&=&-\mathbb{E}_{x_{f} \sim \mathbb{Q}} [\log  (D(x_f) ) ].\nonumber
\end{eqnarray}  
Here, $D(x) = \sigma(C(x))$, where $C(x)$ refers to the final output of the discriminator network after which the activation function $\sigma$ is applied. $x_r$ refers to real data which is sampled from the dataset and $x_f$ refers to data which is generated with generator $G$. 
A more recent GAN framework, Relativistic Average GAN (RaGAN) \cite{RaGAN}, utilizes the following losses instead:
\begin{eqnarray}
L_{RaGAN} \!\!\!&=&\!\!\!-\mathbb{E}_{x_{1}}  [\log   (\overline{D}(x_{1}, x_{2}  )  )  ]-\mathbb{E}_{x_{2}}  [\log   (1- \overline{D} (x_{2}, x_{1}  )  )  ] \nonumber \\
L_{RaGAN}^{D}\!\!\! &=& \!\!\! L_{RaGAN}(x_f,x_r) ,~~ L_{RaGAN}^{G} =  L_{RaGAN}(x_r,x_f) \nonumber
\end{eqnarray}  
where $\overline{D}(x_1,x_2) \triangleq \sigma ( C(x_1) - \mathbb{E}[ C(x_2)])$.  While a discriminator in vanilla GAN predicts how realistic an image is, relativistic discriminator evaluates the realness of real and fake images relatively. As it has been shown that using RaGAN loss provides sharper details while having more stable training, we also incorporate RaGAN loss in our model.
\subsection{PanColorization GAN (PanColorGAN) Model}

\begin{figure*}[th]
\centering
\includegraphics[width=12cm]{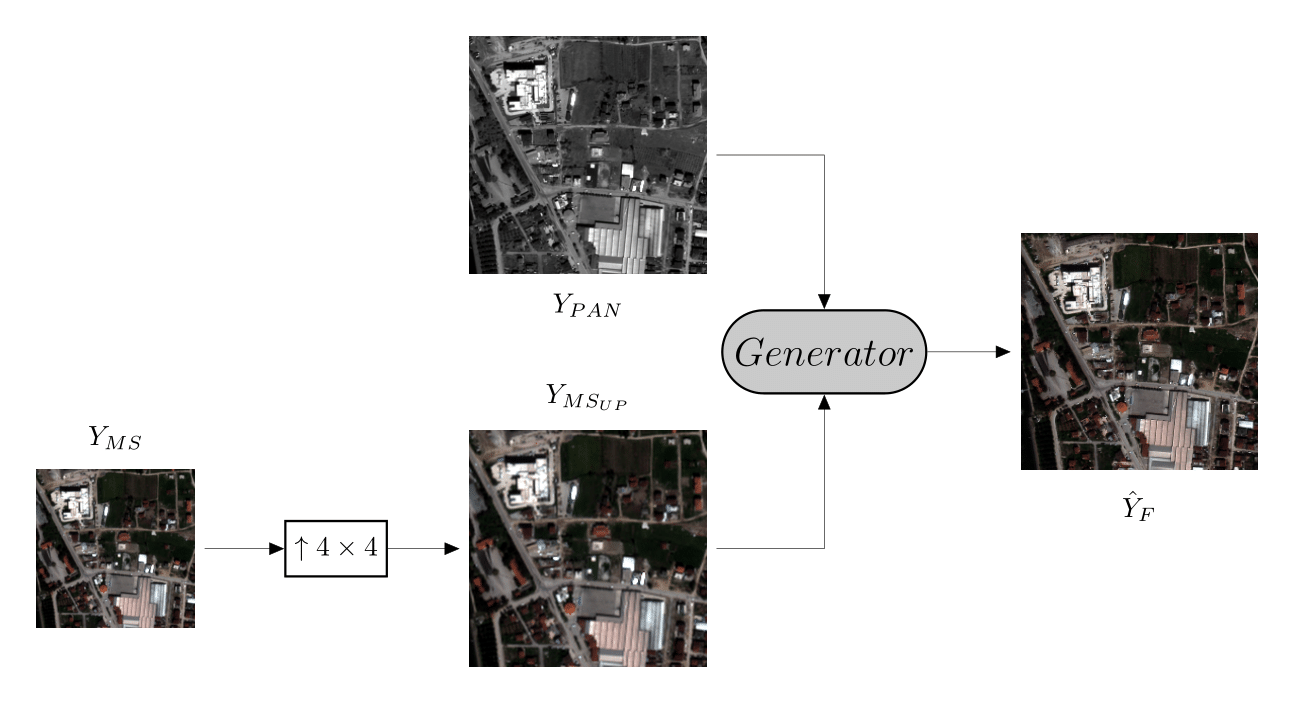}
\caption{Full Resolution Inference (Testing) Scheme: $Y_{MS}$ is upsampled by $4\times$ to obtain $Y_{MS_{UP}}$. $Y_{MS_{UP}}$ and $Y_{PAN}$ are fed to the trained PanColorGAN generator in order to get the full resolution pansharpened image $\hat{Y}_F$ at the output.}
\label{fig:PanColorGANFull}
\end{figure*}

 Figure~\ref{fig:PanColorGANArch}. depicts the details of the PanColorGAN architecture. Its generator $G$ is a modified and expanded version of the UNet \cite{Unet} architecture. It has shortcuts of concatenation across layers in order to provide improved optimization in terms of reducing the vanishing gradients problem. $G$ has four main parts that serve specific goals: (i) spatial detail extraction, (ii) color injection, (iii) feature transformation, and (iv) pansharpened image synthesis. The spatial detail extraction part takes a grayscale image ($X_{GMS}$) as input and applies $3\times3$ convolutions while obtaining color features from the color injection part. The color injection part is a fully convolutional architecture that applies $3\times3$ convolutions four times and injects extracted color features from the multispectral image ($X_{MS}$) to spatial detail extraction layers of the network after every convolution operation except the first one. There is a residual block in the middle of the network that transforms concatenated spatial and spectral features and prepares them for a synthesis of the pansharpened image. Finally, the network slowly increases height and width, and decreases the depth of features by applying upsampling and $3\times3$ convolutions while obtaining features from the detail extraction part, as in the standard Unet architecture. Batch normalization and LeakyReLU activation are inserted after every convolution operation. After obtaining features as the same dimension as the multispectral image, the $\tanh$ activation is applied to map the image intensities to [-1,1] interval.  Using $\tanh$ provides faster and more stable training of GANs \cite{DCGAN}. This produces the output $\hat{Y}_G$ of the generator network.

PanColorGAN discriminator $D$ has a conditional patchGAN architecture \cite{Pix2pix-patchgan}, which operates on image patches, and gives an output for every receptive field it sees. Hence, the output indicates whether those receptive fields seen by $D$ look realistic or not. Then those outputs are aggregated in a patchGAN loss for the discriminator network $D$. The reconstruction loss for the Generator is not calculated over patches, but calculated pixelwise for the whole image. 
In a conditional GAN framework, conventionally the $D$ network takes the output of the  generator or ground truth along with inputs. Pansharpening can be regarded in the framework of  image-to-image translation idea, which was first presented in the study of Isola et al \cite{Pix2pix-patchgan}. In image-to-image translation with conditional adversarial networks, for the discriminator network $D$, the inputs to the generator are taken as conditions in its decision of “real” or “fake”. For that reason, during our training, fake batches consist of grayscale images $X_{GMS}$, reduced multispectral images $X_{MS}$, and outputs of $G$ network $\hat{Y}_G$. Real batches consist of $X_{GMS}$, $X_{MS}$ and original multispectral images $Y_{MS}$. This procedure differs from that of the unconditional generative adversarial networks where the generator network synthesizes images from randomly sampled latent variables and the discriminator receives only the generated images and real images at its input. Providing all related inputs with generated and real images ensure that the discriminator network understands visual relations between input and output images. $D$ applies $4\times4$ convolutions with 2-strides 5 times and reduces height and width while increasing depth. Then a final convolution reduces the depth to 1. Batch normalization and LeakyReLU activation are executed after every convolution layer. Sigmoid operation is applied in order to shrink the interval to [0,1]. Hence, at the output of $D$, indicators of the realness of receptive fields in the given image are obtained.

PanColorGAN model utilizes the following losses for learning the weights of the $G$ and $D$ networks:
\begin{eqnarray}
L_{D} &=& L_{RaGAN}(Y_{MS},\hat{Y}_G)\label{eq:Dis} \\
L_{G} &=& L_{Rec} + \alpha L_{RaGAN}(\hat{Y}_G, Y_{MS}) \label{eq:Gen}\\
L_{Rec} &=&  \parallel Y_{MS} - \hat{Y}_G \parallel_1
\end{eqnarray}  
The reconstruction loss is designed as the Mean Absolute Error (L1 Loss), whereas the adversarial loss is designed as the relativistic average GAN loss. While the reconstruction loss increases pixelwise similarity between generated images and corresponding multispectral images, the adversarial loss brings closer the distribution of generated images to multispectral images and provides sharpness in detail. 
 In PanColorGAN, $L_{Rec}$ measures the distance between $Y_{MS}$ and $\hat{Y_G}$ rather than $\hat{Y_P}$, because the latter would lead the training network to bias the spatial distribution of the pansharpened image towards the multispectral image domain, which is not desirable, as argued before in Section~\ref{sec:Motivation}.

\subsection{Random Downsampling of Multispectral Images}

As we discussed in Section~\ref{subsec:FullResProblems}, training the pansharpening network with $4 \times$ downsampling scale reduces the representation capacity of the model, particularly for the full resolution pansharpening scenario. Hence, we substitute $4 \times$ downsampling operation with a random downsampling operation in an enhanced model, which we call PanColorGAN+RD (Random Downsampling). As we want the model to learn the colorization of grayscale transformed multispectral images and panchromatic images, the model should be robust to variations in the spatial resolutions of the reduced multispectral images, which are  used for their spectral information. When random downsampling procedure is used for an image, say with height and width sizes of 256, instead of downsampling the image to a fixed size of $64 \times 64$, we sample an integer, say $s$, from a uniform random distribution between $(a,b)$, where $a$ and $b$ are two predefined numbers (See Section IV-A). We downsample the image to the selected size $s \times s$, and then immediately upsample it back to $256 \times 256$. We emphasize here that this random downsampling process is applied only during the training phase of the network. In the testing phase, random downsampling is not utilized. This modification provides a way to PanColorGAN to improve its learning as follows: when only $4 \times$ downsampling is used in the training stage, the network learns to interpolate the reduced panchromatic image and the reduced multispectral images with the given scale and does not learn the colorization task properly. As the actual spatial resolution scale difference between the former two is not known exactly, the learned result provides neither the desired nor the sufficient super-resolution level when the model is applied on full resolution. This effect is demonstrated in Section~\ref{subsec:FullRes}.

\subsection{Inference through proposed PanColorGAN models}
After the training phase is completed, during the reduced resolution testing phase, the original $Y_{PAN}$ image is reduced to the same size as the multispectral image to obtain $X_{PAN}$. The $X_{PAN}$ and $X_{MS}$ images are provided to the trained PanColorGAN generator network $G$ and $\hat{Y} = G(X_{PAN},X_{MS})$ is obtained as the output, for the reduced resolution inference. 

Figure~\ref{fig:PanColorGANFull} illustrates how to execute the full resolution, i.e. the real life scenario in pansharpening. The original $Y_{PAN}$ and $4 \times$ upsampled version of $Y_{MS}$ are provided to the trained PanColorGAN generator network $G$, and $\hat{Y}_F = G(Y_{PAN},Y_{MS_{UP}})$ is obtained as the full-resolution pansharpened image output.

\begin{figure*}[th]
\centering
\includegraphics[width=15cm]{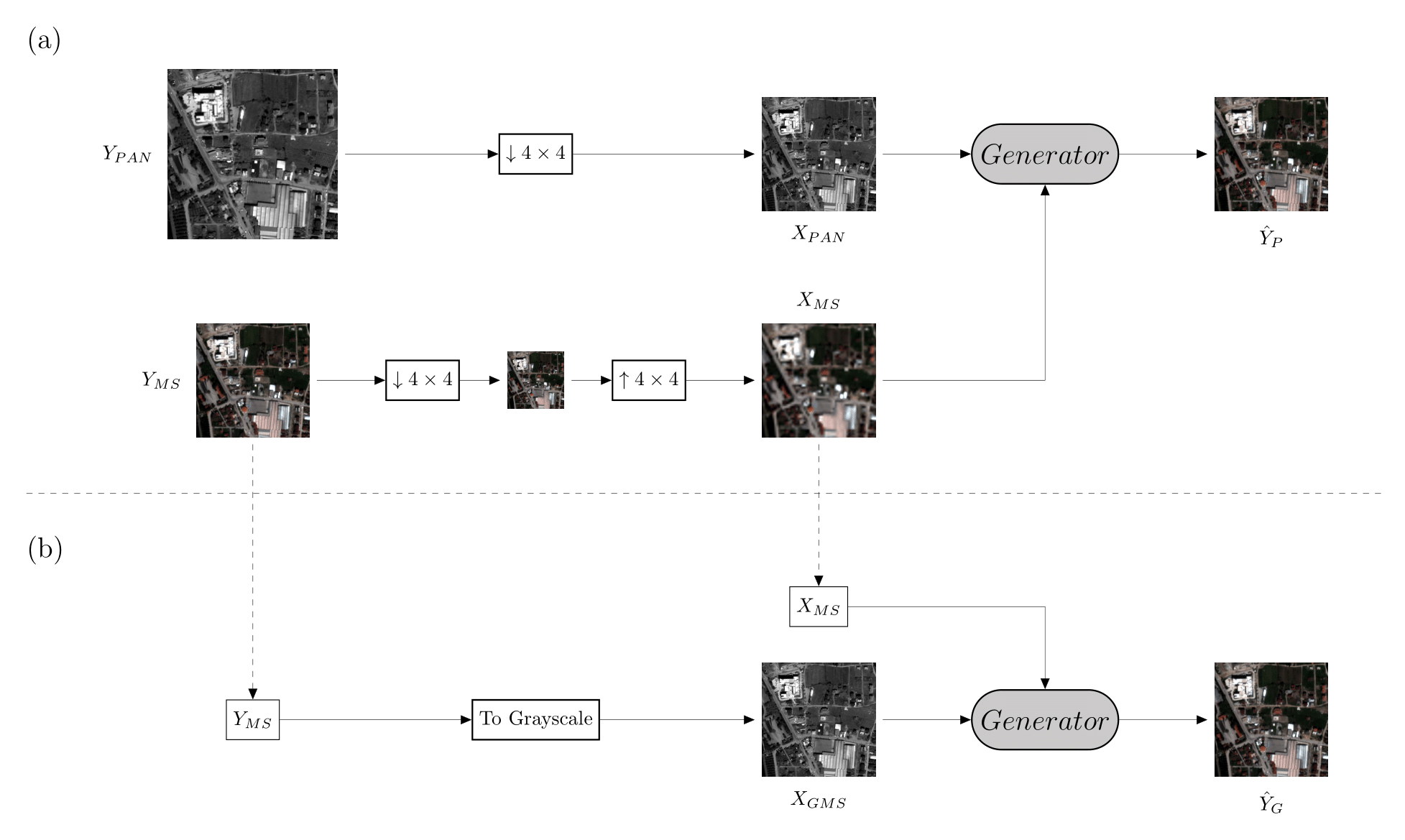}
\caption{Reduced Resolution Testing Scheme. (a) $Y_{PAN}$ is downsampled by $4 \times$ to obtain $X_{PAN}$. $Y_{MS}$ is downsampled and then upsampled by $4 \times$ to obtain the $X_{MS}$ image. $X_{MS}$ and $X_{PAN}$ are given to the generator $G$ in order to get pansharpened image $\hat{Y}_P$ in the natural operation mode of the PanColorGAN, PanSRGAN, and other CNN-based pansharpening models. (b) This mode is shown only for evaluation of the training procedure of PanColorGAN-GMS and PanColorGAN+RD-GMS models: $Y_{MS}$ is converted to grayscale to obtain $X_{GMS}$. $X_{MS}$ and $X_{GMS}$ are fed to the generator $G$ to obtain the colorized image $\hat{Y}_G$.}
\label{fig:PanColorGANTest}
\end{figure*}
\section{Experiments and Results}
In this section, we provide implementation details of experiments, utilized datasets and evaluation indexes, quantitative and qualitative evaluation of reduced resolution and full resolution results. Furthermore, we present transferability properties of our model as well as discussions of the results. For further visual results, we refer to the following website. \footnote{\href{http://vision.itu.edu.tr/supplimentaryceliketal/}{http://vision.itu.edu.tr/supplimentaryceliketal/}}

\label{sec:Results}
\subsection{Implementation Details}
We implemented PanColorGAN in Pytorch 1.0 and trained it on one Titan RTX GPU. An iteration in the training phase takes approximately 2 seconds, which makes an epoch approximately 1 hour for our training set. We trained our models for 100 epochs and selected the best checkpoint in the latest epochs in terms of performance, which took a model 4 days to train. As a baseline GAN-based pansharpening method, we build a pansharpening model inspired by the super-resolution task, which is similar to other standard CNN-based methods. We name it as PanSRGAN, which is trained with $X_{PAN}$ input instead of $X_{GMS}$, following the same procedure in standard CNN-based pansharpening framework. We compare it with our PanColorGAN model in order to perform an ablation study to assess the provided improvements.

Disabling the adversarial loss and using only the reconstruction loss leads to blurrier image generation. This blurriness property occurs due to characteristics of reconstruction loss, for instance as in pixel-wise minimum squared error loss that tends to average details of local neighborhoods. Adversarial loss provides a perceptual similarity metric to training which leads to sharper results in contrast to reconstruction loss \cite{PerceptualSim}. The advantages of using generative adversarial networks instead of only generators with reconstruction loss were reported in the study of Liu et al \cite{PSGAN} for the pansharpening case as well. Considering the beneficial effects of adversarial loss in image generation, we also adapt the generative adversarial network framework to all pansharpening models proposed in this work.

In our experiments, the mini-batch size was set to 16. We used Adam optimizer with an initial learning rate 0.0002, $\beta_1$ as 0.5 and $\beta_2$ as 0.999. We did not use weight decay because it decreased the performance of image synthesis. Adversarial loss weight $\alpha$ was set to 0.005 in Eq.~\ref{eq:Gen}. A leakyReLU activation with 0.2 slope is used in all activation layers. During the training of the PanColorGAN+RD model, for each image in a given batch, a random downsampling size is sampled uniformly as an integer from the $[20, 80]$ interval. The upsampling scale is then automatically set to upscale the downsampled image back to $256$. Both upsampling and downsampling are carried out with a bicubic interpolation scheme.

\vspace*{-0.2cm}
\subsection{Dataset and Evaluation Indexes}

\begin{table*}[!t]
\renewcommand{\arraystretch}{1.3}
\caption{INFORMATION OF SATELLITE IMAGES IN DATASETS}\label{satellite_table}
\centering
\begin{tabular}{|c| c| c| c| c| c| c|}
\hline
 Region & Satellite & Acquisition Date & MS/PAN m & Across Track & Along Track & Train/Test \\
  \hline
 Aydin & Pleiades 1A & 3/3/2017 & 2 / 0.5 & -6.91 & 18.12 & Train \\
 Istanbul & Pleiades 1A & 4/29/2017 & 2 / 0.5 & -22.92 & -11.15 & Train \\
 Istanbul & Pleiades 1A & 11/25/2017 & 2 / 0.5 & 4 & -18.77 & Train \\
 Bursa & Pleiades 1A & 4/4/2018 & 2 / 0.5 &  4.32 & -14.60 & Train \\
 Bilecik & Pleiades 1A & 4/25/2017 & 2 / 0.5 & 3.08 & -13.89 & Train \\
 Mugla & Pleiades 1A & 2/6/2017 & 2 / 0.5 & -9.08 & 15.73 & Test \\
 \hline
 Stockholm & Worldview 2 & 8/27/2016 & 1.6 / 0.4 & 6.20 & -7.10 & Test \\
 Rio & Worldview 3 & 5/2/2016 & 1.2 / 0.3 & 23.90 & -2.50 & Test \\
 Tripoli & Worldview 3 & 3/8/2016 & 1.2 / 0.3 & -3.70 & 5.00 & Test \\
 Washington & Worldview 2 & 8/15/2016 & 1.6 / 0.4 & 10.10 & -7.70 & Test \\
\hline
\end{tabular}
\end{table*}

The first dataset consists of 6 full-sized image scenes from Pleiades 1A\&1B twin satellites owned by AIRBUS. Five of them are used for training and one of them is used for testing. Frames are divided into patches of $1024\times 1024$ for panchromatic images, $256\times 256$ for multispectral images. Thus, 30000 training samples and 5700 test samples are gathered for the Pleiades dataset. Pleiades image data includes 4 channels for multispectral images which are red, green, blue, and near-infrared with $2m$ resolution. Its single-banded panchromatic image has $0.5m$ resolution. The dataset consists of images from both rural and urban areas in Turkey. In addition, image acquisition angles and seasons are in a wide range, which helps to train the model with a dataset that reflects different illumination and geometric conditions. The second dataset we utilize in our testing experiments consists of four image scenes from Worldview 2 and Worldview 3 satellites owned by Digital Globe (Maxar Technologies), which is published as open source \cite{DigitalGlobe}. We extract 350 patches ($256\times 256$ MS, $1024\times 1024$ PAN) from 4 cities, which are Stockholm, Washington, Tripoli and Rio. Similar to the Pleiades dataset, Digital Globe data has 4 channels for multispectral images which are red, green, blue, and near-infrared. The spatial resolution of 4-band multispectral data is 1.6m and single panchromatic data is 0.4m for Worldview 2 images, while the resolution of 4-band multispectral data is 1.2m and single panchromatic data is 0.3m for Worldview 3 images. Both Pleiades and Worldview images were obtained in UTM projection system with appropriate zones. Detailed information about the image dataset is provided in Table ~\ref{satellite_table}. We trained the following models: (i) the proposed PanColorGAN; (ii) PanColorGAN+RD: PanColorGAN with Random Downsampling; (iii) PanSRGAN: the baseline GAN-based pansharpening model; (iv) TA-CNN: Target-Adaptive CNN-based pansharpening \cite{TargetAdaptiveCNN}; (v) PanNet: A Deep Network Architecture for Pan-Sharpening \cite{PanNet}. For comparison, we also utilize traditional pansharpening algorithms that are available in the Open Remote Sensing repository \cite{OpenRemoteSensing} 
which are BDSD \cite{BDSD}, ATWT \cite{ATWT}, GSA \cite{GSA}, GLP-REG-FS \cite{GLPREGFS}, NIHS \cite{NonlinearIHS}, and Semiblind Deconv \cite{SemiblindDeconv}. For training TA-CNN and PanNet models, we used the codes provided by the authors \cite{TargetAdaptiveCNN_Site,PanNet_Site}.

For the quantitative analysis, across all algorithms including the baselines, QAVE \cite{Q}, SAM \cite{SAM}, ERGAS \cite{ERGAS}, sCC \cite{sCC}, and Q \cite{Q4} are used as performance measures that include references in their calculations. We also analyze all algorithms in full resolution with no-reference metrics. Non-reference performance measures we utilize are $D_{s}$, $D_{\lambda}$, and QNR \cite{QNR}. For calculation of all metrics, again we use the Matlab\texttrademark code in Open Remote Sensing repository \cite{OpenRemoteSensing}.

\vspace*{-0.15cm}
\subsection{Evaluation of Reduced Resolution Results}

Figure~\ref{fig:PanColorGANTest} depicts reduced resolution testing scheme. We construct two versions of the method during inference, where we provide: (1) grayscale multispectral image alongside reduced multispectral images to obtain PanColorGAN-GMS; (2) reduced panchromatic image alongside reduced multispectral images to obtain PanColorGAN-PAN model. Similarly, two versions PanColorGAN+RD-GMS and PanColorGAN+RD-PAN models are constructed for the random-downsample version of our method. Reduced panchromatic images and reduced multispectral images are utilized for traditional pansharpening algorithms, CNN-based methods, and the PanSRGAN model.

\subsubsection{Quantitative Analysis of Reduced Resolution Results}
For all the with-reference measures in Table~\ref{reduced_uhuzam_table}, PanColorGAN-GMS outperformed all other techniques, both CNN-learning based, and previous traditional approaches. PanColorGAN-GMS surpasses PanColorGAN-PAN extension models, where for the latter, the reduced panchromatic image is used as the input during inference. This is expected because the training procedure is set up to force the model to learn to colorize the gray-transformed multispectral image, hence the loss functions make use of the grayscaled multispectral images, not the reduced panchromatic images. Also, although standard CNN-based models such as PanNet, TA-CNN, and PanSRGAN perform clearly worse in visual quality (demonstrated later), they obtain second-tier yet close performances to other PanColorGANs, still staying behind PanColorGAN-GMS.

\begin{table}[!t]
\renewcommand{\arraystretch}{1.3}
\caption{WITH-REFERENCE PERFORMANCE INDICATORS AT REDUCED RESOLUTION ON
PLEIADES DATASET}\label{reduced_uhuzam_table}
\centering
\begin{tabular}{l c c c c c}
\hline
 & QAVE & Q & sCC & SAM & ERGAS \\
 (worst-best) & (0-1) & (0-1) & (0-1) & (inf-0) & (inf-0) \\
  \hline
  BDSD & .692 & .673 & .792 & 2.649 & 3.049 \\
  ATWT & .718 & .704 & .780 & 2.226 & 2.669 \\
  GSA & .689 & .669 & .774 & 2.535 & 3.177 \\
  GLP-REG-FS & .716 & .702 & .795 & 2.329 & 2.815 \\
  Nonlinear IHS & .698 & .682 &.821 & 1.873 & 2.597 \\
  Semi-blind Convolution & .712 & .700 & .750 & 2.276 & 19.179 \\
PanNet & .885 & .882 & .911 & 1.803 & 1.440 \\
TA-CNN & \textcolor{blue}{\textbf{.891}} & \textcolor{blue}{\textbf{.888}} & \textcolor{blue}{\textbf{.933}} & \textcolor{blue}{\textbf{1.509}} & \textcolor{blue}{\textbf{1.295}} \\
\hline
PanSRGAN & .917 & .889 & .960 & 1.759 & 1.480 \\
PanColorGAN-GMS & \textcolor{red}{\textbf{.956}}  & \textcolor{red}{\textbf{.942}} & \textcolor{red}{\textbf{.981}} & \textcolor{red}{\textbf{1.362}} & \textcolor{red}{\textbf{1.039}} \\
PanColorGAN-PAN & .808 & .780 & .857 & 2.116 & 2.222 \\
PanColorGAN+RD-GMS & .949 & .930 & .976 & 1.620 & 1.219 \\
PanColorGAN+RD-PAN & .794 & .763 & .850 & 2.351 & 2.447 \\

\hline
\end{tabular}
\end{table}

\subsubsection{Reduced Resolution Scenario Visual Results}

\begin{figure*}[ht]
\centering
\includegraphics[width=16cm]{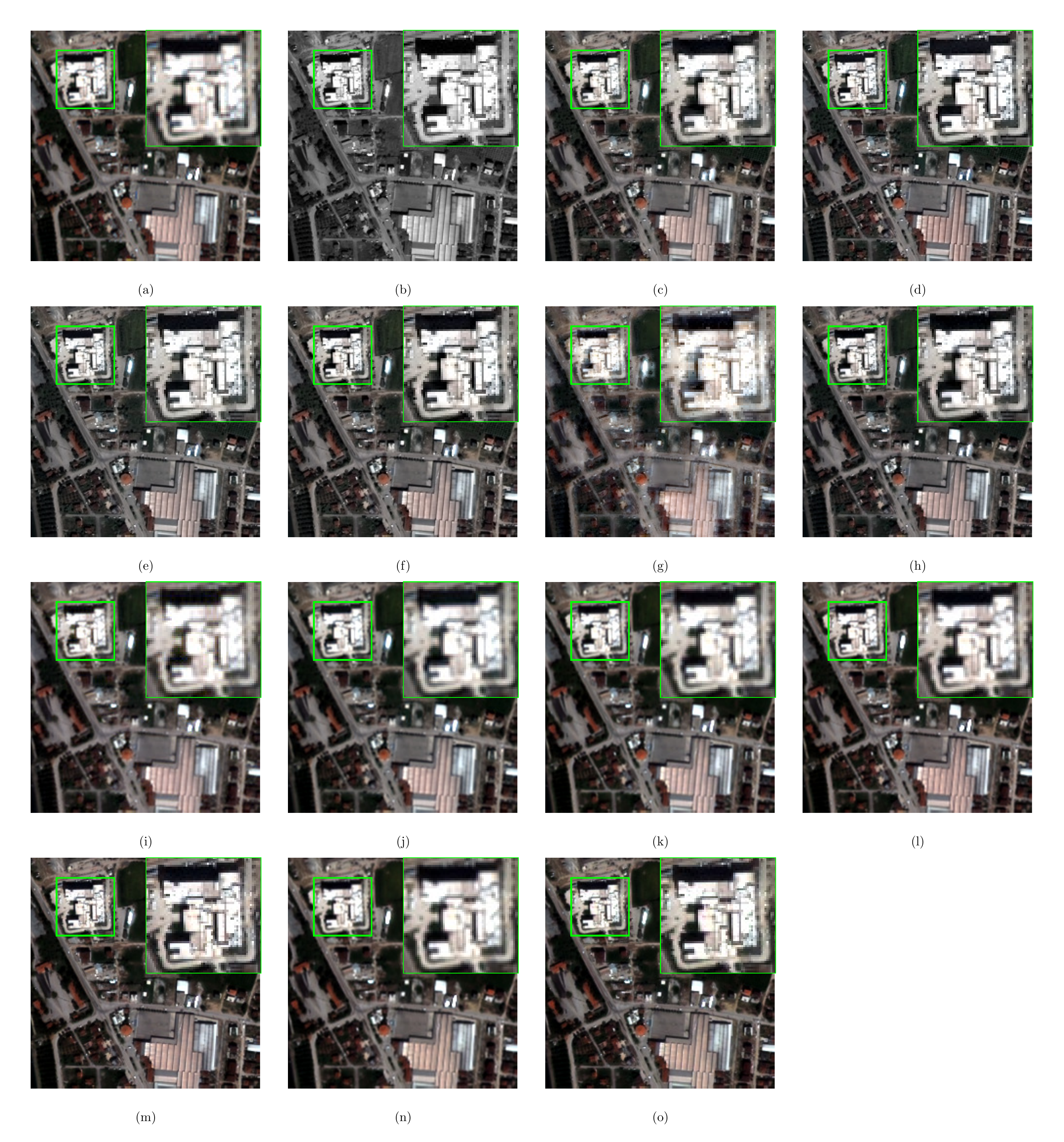} 
\caption{Reduced Resolution Scheme Test Results for Baseline methods and PanColorGAN models over Pleiades Dataset: (a) Multispectral (b) Reduced Resolution Panchromatic (c) ATWT (d) BDSD (e) GSA (f) GLP-REG-FS (g) Nonlinear IHS (h) Semi-blind Convolution  (i) TA-CNN (j) PanNet (k) PanSRGAN (l) PanColorGAN-GMS (m) PanColorGAN-PAN (n) PanColorGAN+RD-GMS  (o) PanColorGAN+RD-PAN. Region in green box in each picture is zoomed and pasted on the top right for visualization.}
\label{reduced_uhuzam_images}
\end{figure*}

Figure~\ref{reduced_uhuzam_images} shows results from all algorithms on Pleiades test dataset. The corresponding full-resolution panchromatic image was given in Figure~\ref{fig:sharpness} on the left. Images in (c)-(h) belong to the results of traditional approaches, and (i)-(o) depict results of the CNN-based methods. Artifacts in the Nonlinear IHS in (g) are immediately noticeable. The continuity in lines, as well as sharp contrast changes across regions of the pinkish roofs of an industrial complex in the bottom center parts of the image, is preserved only in a few methods. Among those, PanColorGAN-PAN (m) reproduced those features most successfully, followed by BDSD (d), PanColorGAN+RD-PAN (o).  Similarly, the spectral or the color reproduction in the results can be gauged from the orange rooftops. Those colors are preserved best in all PanColorGAN models, and PanSRGAN to a degree, whereas the traditional methods all lack the color saturation level of the original multispectral image (a). PanNet (j) and TA-CNN (i) also provided similar visual results to PanSRGAN (k), however, it can be observed that they could not preserve spatial details. The blurring characteristics of the methods are clearly visible, starting with Nonlinear IHS, ATWT, and relatively in all traditional methods except BDSD. Among CNN-based approaches, PanNet, TA-CNN, PanSRGAN, PanColorGAN-GMS, and PanColorGAN+RD-GMS methods show blurrier characteristics with respect to the PanColorGAN-PAN and PanColorGAN+RD-PAN methods, which both clearly outperform all the methods in visual inspection in terms of both structural and spatial properties while keeping spectral properties in an acceptable level when compared to the original multispectral image visually. Although, we obtain higher quantitative scores for PanColorGAN-GMS when compared to the PanColorGAN-PAN variants, it is well-known that higher quantitative scores do not necessarily indicate better perceptual results, as this was also reported in the literature \cite{PNN}.

\subsection{Evaluation of Full Resolution Results}
\label{subsec:FullRes}
We evaluate the quantitative and qualitative results of the full-resolution experiments in this section.

\subsubsection{Quantitative Analysis of Full Resolution Results}

Table~\ref{full_uhuzam_table} refers to calculated performance measures that require no-reference, as a ground truth or reference pansharpened image does not exist in the real-life full-resolution scenario. TA-CNN provides the best quantitative performance among previous methods followed by PanNet, Nonlinear IHS, and BDSD, whereas both PanColorGAN and PanSRGAN achieve similar results. The measure $D_{\lambda}$ focuses on spectral characteristics and $D _{s}$ focuses on spatial details, whereas QNR is a combination of both measures. In spectral measures, PanSRGAN achieves a good performance in $D_{\lambda}$, whereas TA-CNN achieves the best performance in $D _{s}$.

\begin{table}[!t]
\renewcommand{\arraystretch}{1.3}
\caption{NO-REFERENCE PERFORMANCE INDICATORS AT FULL RESOLUTION ON
PLEIADES DATASET}\label{full_uhuzam_table}
\centering
\begin{tabular}{l c c c}
\hline
 & $D_{\lambda}$ & $D _{s}$ & QNR \\
 (worst-best) & (inf-0) & (inf-0) & (0-1)\\
 \hline
BDSD & \textcolor{blue}{\textbf{.037}} & .094 & .872 \\
ATWT & .101 & .178 & .740\\
GSA & .132 & .313 & .598 \\
GLP-REG-FS & .089 & .150 & .774 \\
Nonlinear IHS & .046 & .080 & .876 \\
Semi-blind Convolution & .123 & .227 & .678 \\
PanNet & .060 & .044 & .895 \\
TA-CNN & .041 & \textcolor{blue}{\textbf{.037}} & \textcolor{blue}{\textbf{.920}} \\
\hline
PanSRGAN & \textcolor{red}{\textbf{.015}} & .117 & \textcolor{red}{\textbf{.869}} \\
PanColorGAN & .042 & \textcolor{red}{\textbf{.099}} & .862 \\
PanColorGAN+RD & .048 & .134 & .824 \\
\hline
\end{tabular}
\end{table}

\subsubsection{Full Resolution Scenario Visual Results}

\begin{figure*}[ht]
\centering
\includegraphics[width=16cm]{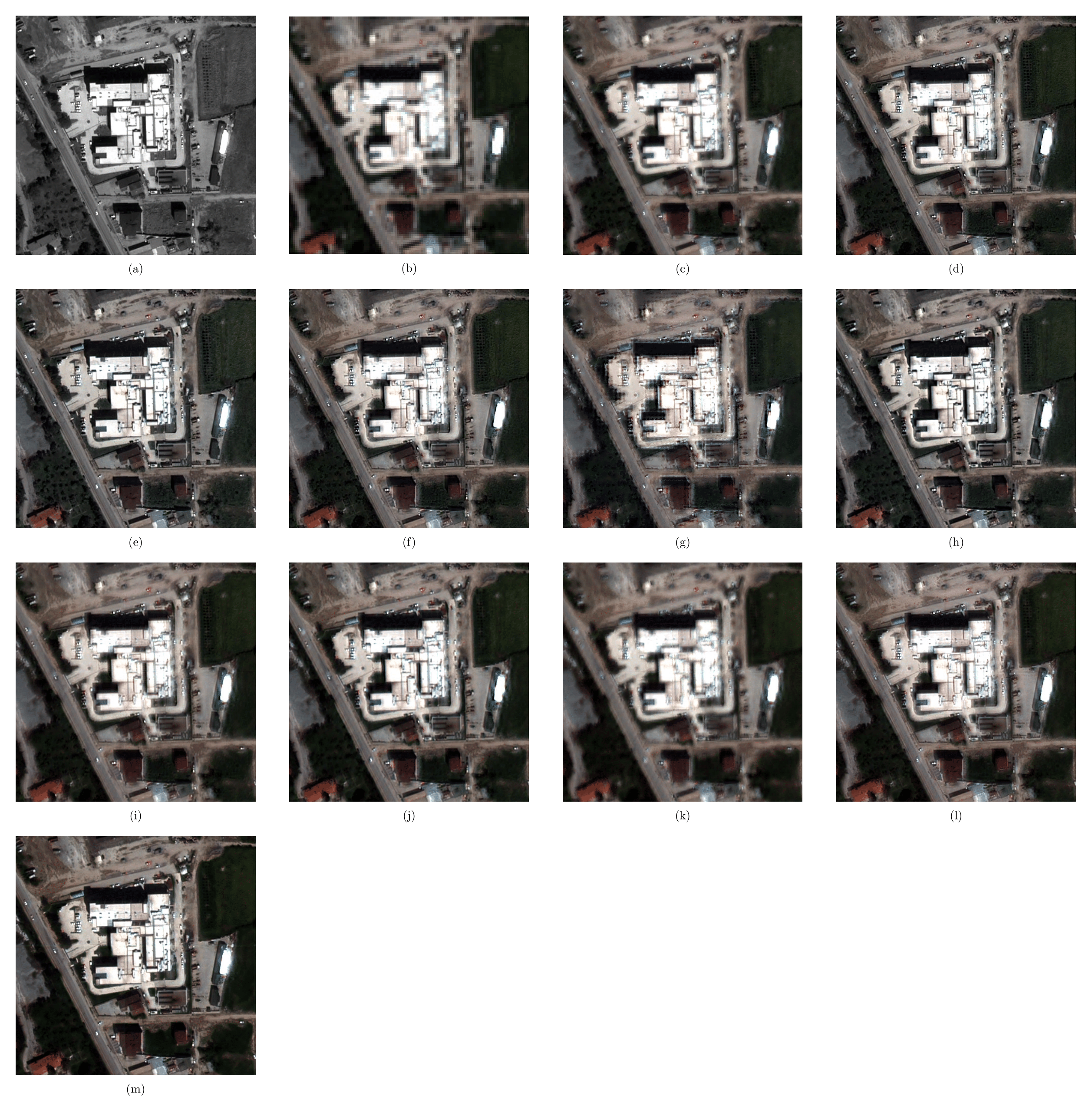}
\caption{Full resolution results for Pleiades test Dataset: (a) Panchromatic (b) Multispectral (c) BDSD (d) ATWT (e) GSA (f) GLP-REG-FS (g) Nonlinear IHS (h) Semi-blind Convolution  (i) TA-CNN (j) PanNet (k) PanSRGAN (l) PanColorGAN (m) PanColorGAN+RD.}
\label{full_uhuzam_images}
\end{figure*}

Figure~\ref{full_uhuzam_images} shows full resolution results from all algorithms on the Pleiades test dataset. Images in (a) and (b) refer to the input, i.e. the original panchromatic and multispectral images, respectively. Images in (c)-(h) refer to results produced by traditional methods, whereas (i)-(m) refer to CNN-based methods. Artifacts in results of BDSD (c), (GLP-REG-FS (f), and Nonlinear IHS (g) from traditional methods, as well as in results of PanSRGAN (k), PanColorGAN (l) are apparently visible. Although PanNet (i) and TA-CNN (j) gave decent results in no-reference metrics, visual results do not support those numbers. They produce more blurry results when they are compared with PanColorGAN+RD. Among the traditional methods, GSA (e) and Semi-blind Convolution (h) produce better results than the former, whereas PanColorGAN+RD (m) provides the best performance. For instance, when the bending corner segments of the white complex structures in the middle of the image are compared, better preservation of continuity of borders is observed in the PanColorGAN+RD method and traditional methods: GSA and Semi-blind Convolution. The sharp edges and high contrast between the white structures and its surroundings is best captured in PanColorGAN+RD and GSA, where the smearing across regions is minimal. In the green fields with tree clusters and vegetation towards top right and bottom left of the scene in the figure, GSA and Semi-blind Convolution preserve the original pattern better than all other methods. One can also observe that due to the low resolution of the MS in (b), the terrain color looks yellow due to the relatively blurry characteristic of the image, whereas the proposed PanColorGAN+RD (m) produces a gray-yellow tone, which matches the colors in other methods. It can be fairly said that all CNN-based techniques are losing the vertical lines of the trees to a degree. This is one limitation we observed in most of the MRA pansharpening methods, including CNN-based methods. In terms of spectral color features, almost all of the techniques including PanColorGANs are observed to capture the original color distributions of the multispectral input image in (b). In terms of spatial features, PanColorGAN+RD shows the best performance, as it includes randomness introduced in its downscaling ratios that increases its robustness to minute resolution variations between the reduced panchromatic and multispectral images.

\subsection{Discussions and Transferability}

\begin{table}[!t]
\renewcommand{\arraystretch}{1.3}
\caption{WITH-REFERENCE PERFORMANCE INDICATORS AT REDUCED RESOLUTION ON
DIGITAL GLOBE DATASET}\label{reduced_dg_table}
\centering
\begin{tabular}{l c c c c c}
\hline
 & QAVE & Q & sCC & SAM & ERGAS \\
 (worst-best) & (0-1) & (0-1) & (0-1) & (inf-0) & (inf-0) \\
  \hline
  BDSD & \textcolor{blue}{\textbf{.832}} & .831 & \textcolor{blue}{\textbf{.833}} & 7.259 & 4.803 \\
  ATWT & .830 & \textcolor{blue}{\textbf{.843}} & .827 & \textcolor{blue}{\textbf{6.110}} & \textcolor{blue}{\textbf{4.628}} \\
  GSA & .814 & .834 & .801 & 7.076 & 4.952 \\
  GLP-REG-FS & .820 & .834 & .807 & 6.798 & 4.777 \\
  Nonlinear IHS & .755 & .754 & .766 & 6.229 & 5.808 \\
  Semi-blind Convolution & .832 & .836 & .813 & 6.062 & 12.219 \\
PanNet & .690 & .681 & .633 & 7.382 & 6.998 \\
TA-CNN & .673 & .665 & .622 & 7.590 & 7.166 \\
\hline
PanSRGAN & .764 & .727 & .792 & 7.785 & 7.430 \\
PanColorGAN-GMS & \textcolor{red}{\textbf{.884}} & \textcolor{red}{\textbf{.845}} & \textcolor{red}{\textbf{.936}} & \textcolor{red}{\textbf{6.783}} & \textcolor{red}{\textbf{4.707}} \\
PanColorGAN-PAN & .835 & .796 & .879 & 9.095 & 6.789 \\
PanColorGAN+RD-GMS & .863 & .828 & .930 & 7.746 & 5.131 \\
PanColorGAN+RD-PAN & .813 & .776 & .857 & 9.319 & 7.182 \\

\hline
\end{tabular}
\end{table}

\begin{table}[!t]
\renewcommand{\arraystretch}{1.3}
\caption{NO-REFERENCE PERFORMANCE INDICATORS AT FULL RESOLUTION ON
DIGITAL GLOBE DATASET}\label{full_dg_table}
\centering
\begin{tabular}{l c c c}
\hline
 & $D_{\lambda}$ & $D _{s}$ & QNR \\
 (worst-best) & (inf-0) & (inf-0) & (0-1)\\
 \hline
BDSD & .057 & .061 & .886 \\
ATWT & .091 & .146 & .777\\
GSA & .078 & .160 & .775 \\
GLP-REG-FS & .084 & .141 & .788 \\
Nonlinear IHS & \textcolor{blue}{\textbf{.036}} & \textcolor{blue}{\textbf{.046}} & \textcolor{blue}{\textbf{.919}} \\
Semi-blind Convolution & .089 & .131 & .792 \\
PanNet & .041 & .051 & .909 \\
TA-CNN & .062 & .067 & .874 \\
\hline
PanSRGAN & \textcolor{red}{\textbf{.027}} & \textcolor{red}{\textbf{.043}} &\textcolor{red}{\textbf{.930}}  \\
PanColorGAN & .040 & .073 & .890 \\
PanColorGAN+RD & .061 & .070 & .874 \\
\hline
\end{tabular}
\end{table}

Next, we discuss the transferability capability of the PanColorGAN models, as well as all other baseline methods. For that purpose, the trained CNN-based models on the Pleiades dataset are directly tested on the Digital Globe data in order to assess the transferability of the methods. In Table~\ref{reduced_dg_table}, with-reference performance measures for the Digital Globe dataset are given. Again, as in Table~\ref{reduced_uhuzam_table}, PanColorGAN-GMS outperformed all other techniques, including traditional methods. 
PanNet and TA-CNN trained on Pleiades Dataset could not provide satisfactory results when they are tested with Digital Globe Dataset which involves different sensor settings.
Due to different spatial and spectral resolution characteristics of Pleiades and Digital Globe datasets, a slight decrease in all the quantitative measures are naturally observed for CNN-based methods. Yet PanColorGAN models maintain a slighter decrease when they are compared to other CNN-based methods which are PanNet, TA-CNN, and PanSRGAN. Table~\ref{full_dg_table} refers to no-reference performance measures in the full-resolution mode. Nonlinear IHS achieves the best scores among traditional methods, and PanSRGAN gets the highest scores for the three measures.

The real-life pansharpening application with the full-resolution generation deserves further discussions. It is interesting to note that although Nonlinear IHS gives the best quantitative performance with no-reference measures among traditional methods (Table~\ref{full_dg_table}), it was clearly observed that it performed almost the worst in visual inspection in Figure~\ref{full_uhuzam_images}. This experiment highlighted the unreliability and mismatch of the no-reference measures against human visual perception. This finding was also reported by Vivone et al. where many pansharpening algorithms are compared \cite{ComparisonSurvey}. Therefore, in the full-resolution mode, a more reliable evaluation is carried out by visual inspection rather than no-reference quantitative scores.  

\begin{figure*}[ht]
\centering
\includegraphics[width=16cm]{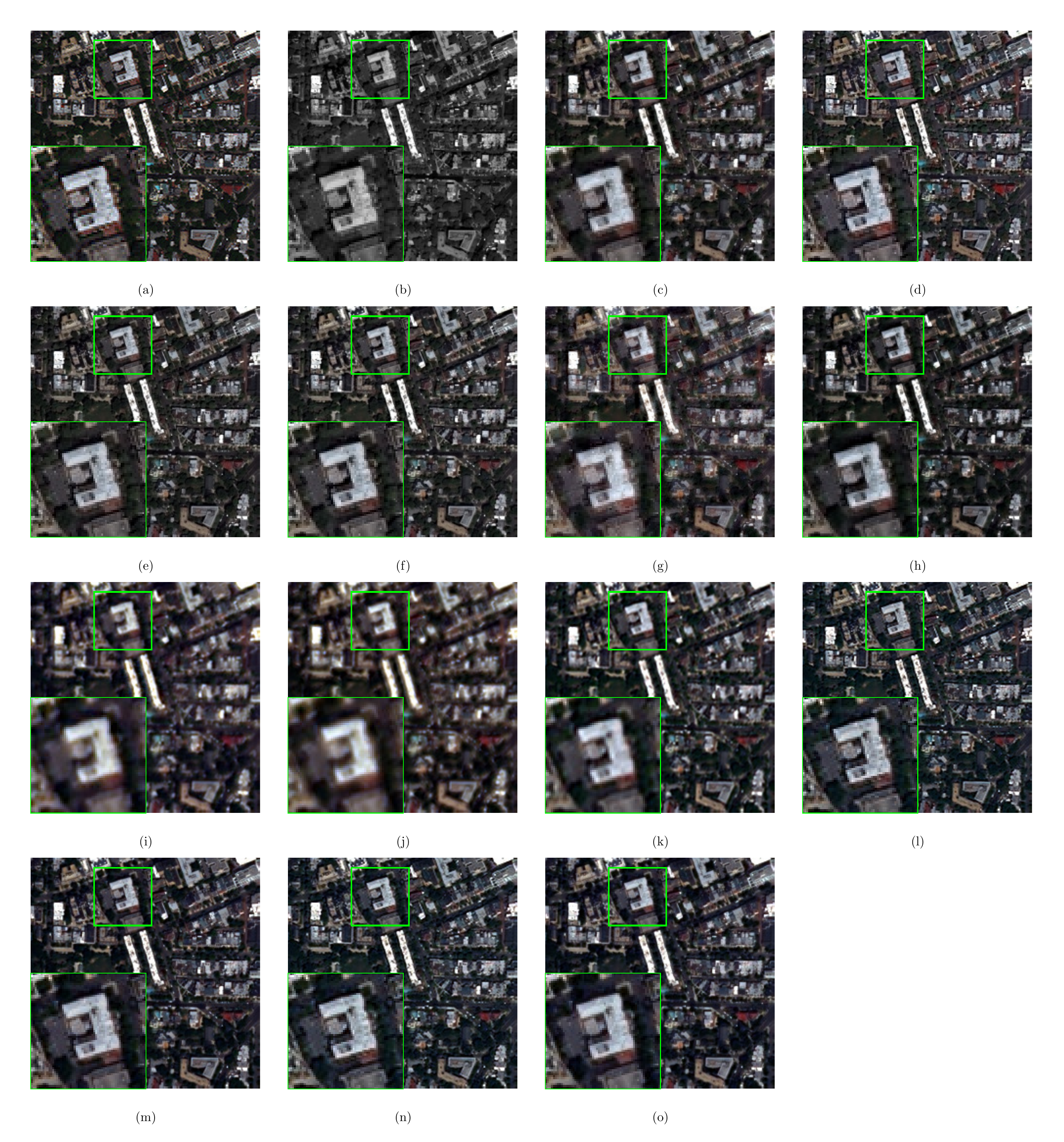}
\caption{Reduced Resolution Scheme Test Results for Baseline methods and PanColorGAN models for Digital Globe Dataset: (a) Multispectral (b) Reduced Resolution Panchromatic (c) ATWT (d) BDSD (e) GSA (f) GLP-REG-FS (g) Nonlinear IHS (h) Semi-blind Convolution  (i) TA-CNN (j) PanNet (k) PanSRGAN (l) PanColorGAN-GMS (m) PanColorGAN-PAN (n) PanColorGAN+RD-GMS  (o) PanColorGAN+RD-PAN. Region in green box in each picture is zoomed and pasted at the bottom for visualization.}
\label{reduced_dg_images}
\end{figure*}

\begin{figure*}[ht]
\centering
\includegraphics[width=16cm]{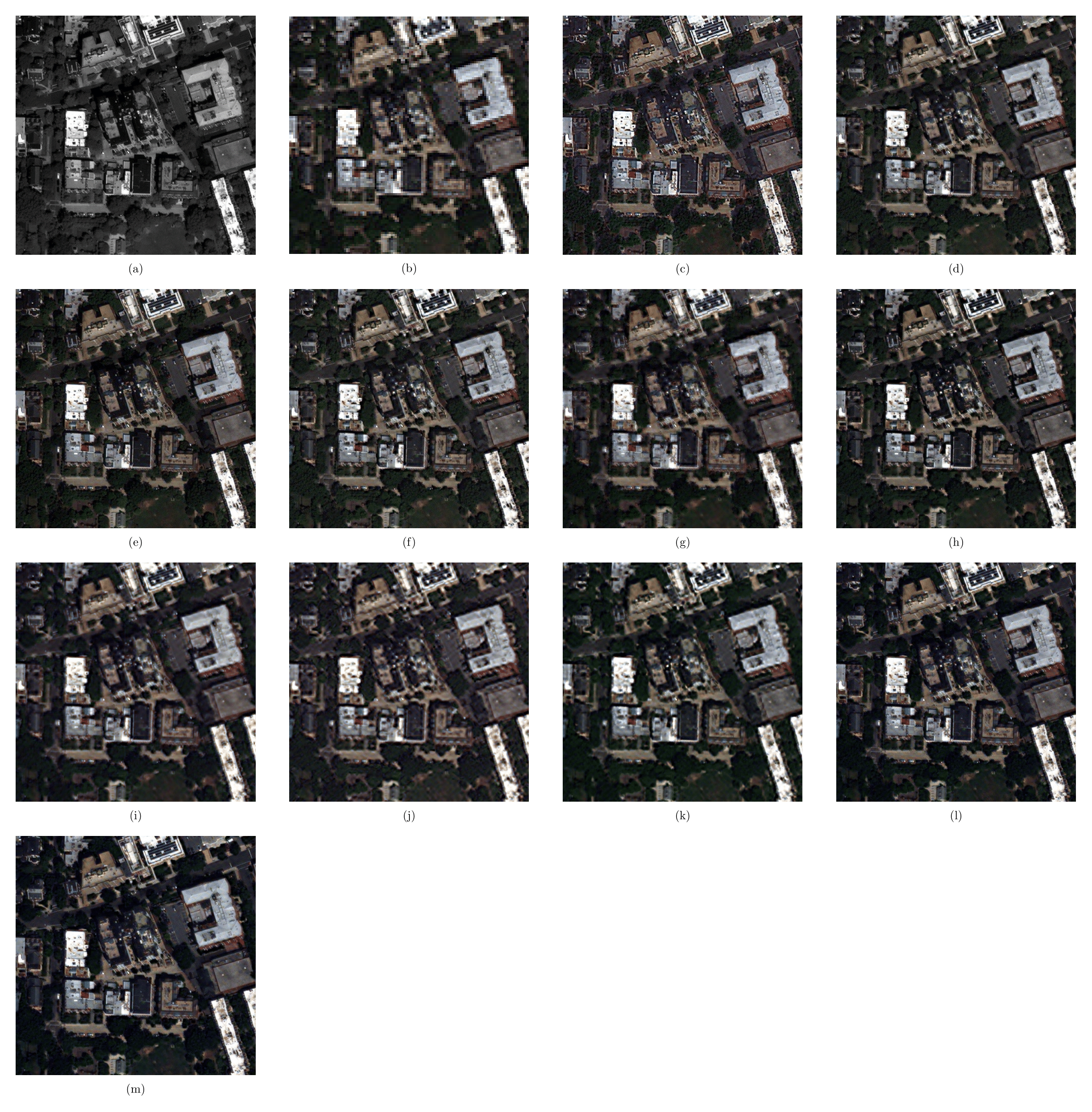}
\caption{Full resolution results for Digital Globe Dataset: (a) Panchromatic (b) Multispectral (c) BDSD (d) ATWT (e) GSA (f) GLP-REG-FS (g) Nonlinear IHS (h) Semi-blind Convolution  (i) TA-CNN (j) PanNet (k) PanSRGAN (l) PanColorGAN (m) PanColorGAN+RD.}
\label{full_dg_images}
\end{figure*}

Figure~\ref{reduced_dg_images} shows visual results from all algorithms in reduced resolution mode on the Digital Globe dataset.  Images in (b)-(h) belong to the results of traditional approaches, and (i)-(o) depict results of the CNN-based methods. This is a heterogeneous image patch with many fine man-made structures and fine textural details. Therefore, the artifacts that were observed with Nonlinear IHS (g) before in Figure~\ref{reduced_uhuzam_images} is not that apparent to the eye. However, the first observation that can be easily made is that results of ATWT (c), GLP-REG-FS (f), Nonlinear IHS (g), PanNet (k), TA-CNN (i) and PanSRGAN (k) present blurrier characteristics than the others. Although we were expecting similar results to PanSRGAN, PanNet and TA-CNN gave slightly worse results in terms of spatial quality in reduced resolution tests. As before, the PanColorGAN models are among the best performers, as can be observed over the fine structures in the zoomed flipped C shaped white building. 
On the other hand, as expected GMS versions of the PanColorGAN provide similar results as the multispectral image while PAN versions preserve spatial details of the reduced panchromatic image. In terms of restoring the color properties, BDSD in (d) and PanColorGAN models (l-o) provides the best visual performance. 

Figure~\ref{full_dg_images} shows visual results from all algorithms in full resolution mode on the Digital Globe dataset. Images in (a) and (b) refer to the input, i.e. the original panchromatic and multispectral images, respectively. Images in (c)-(h) refer to results produced by traditional methods, whereas (i)-(m) refer to CNN-based methods. The lack of preservation for the spectral and spatial properties of the input panchromatic and multispectral images as well as artifacts are clearly visible in BDSD (c), ATWT (d), GLP-REG-FS (f), Nonlinear IHS (g), semi-blind Convolution (h), TA-CNN (i), PanNet (j) and PanSRGAN in (k). We observe that for the Digital Globe dataset, although the problem of spatial detail disagreement between reduced panchromatic and original multispectral images still persists, it is a less pronounced issue compared to the Pleiades dataset, and this is reflected in the closer quantitative performance results between the PanColorGAN and PanColorGAN+RD. However, when full resolution results in Figures \ref{full_uhuzam_images} and \ref{full_dg_images}  are visually inspected, the differences between PanColorGAN and PanColorGAN+RD are clearly observed, where PanColorGAN+RD shows sharper edges and higher contrast than PanColorGAN, which clearly demonstrate the effectiveness of random downsampling in better preservation of spatial details.

A limitation in the development of pansharpening methods is the lack of common datasets. Although standard CNN-based methods, including GAN models, were employed recently for pansharpening, none of those can be evaluated on common data distributions. Naturally, those CNN-based methods were trained and tested on different data distributions, which certainly affects the performance of the models independently from architectural developments. However, our methodological development lies mainly in introduction of a new framework rather than architectural changes, that is why we build a baseline model PanSRGAN with the standard CNN/GAN-based framework, which was crucial to present our improvements in the results.

Our experimental results demonstrate that commonly utilized quantitative image evaluation measures do not necessarily match the expected visual evaluation outcomes. This is not a novel finding, which is also not limited to the domain of satellite imaging. Generally, devising new quantitative image evaluation measures that are faithful to human perceptual evaluations is an open research problem in image analysis.

To summarize our findings, PanColorGAN models are observed to perform at the top among all methods in preserving structural and spatial features of images while keeping the spectral distortion at an acceptable level. This can be asserted for both reduced-resolution and full-resolution modes. In addition, although Digital Globe and Pleiades datasets have different characteristics, PanColorGAN demonstrated better transferability properties than other CNN-based models, as evidenced both quantitatively and qualitatively in our experiments.

\section{Conclusion}
\label{sec:Conc}

We presented a novel pansharpening framework based on GANs and a guided colorization task for coloring the gray-transformed multispectral images. PanColorGAN model, which is positioned on this framework along with two new developments, namely the color injection and the random scale downsampling, demonstrated improved structural preservation and reduced blurring effects when compared to previous CNN-based pansharpening models. The PanColorGAN demonstrates the current state-of-the-art performance both in reduced-resolution and full-resolution pansharpening models especially through visual inspection. It also presents better transferability between different satellite images.

PanColorGAN achieves excellent spatial detail preservation, while the spectral information injection efficiency is open to improvement. Finding ways to preserve the spatial and spectral properties in a balanced manner remains an open future research direction in the problem of pansharpening. We further articulate that the new deep learning-based pan-sharpening methods should elaborate extensively on the full-resolution mode results and transferability, as they certainly present the real challenges in pansharpening.

Pansharpened images produced by the PanColorGAN model can be used effectively in engineering applications such as object detection tasks and mapping purposes such as vector data production (digitization), where high spatial quality and accurate object geometry are required. As spectral properties are also preserved at an acceptable level, the use of these images in satellite image classification tasks is also promising, however, it needs further evaluation.  As for future extensions of this work, the integration and performance evaluations of the medium spatial resolution satellite images with a higher number of multispectral bands and different MS/PAN ratios such as Landsat 8 OLI are planned.


%

\appendices

\section*{Acknowledgments}
\addcontentsline{toc}{section}{Acknowledgment}
This work was supported by Research Fund of the
Istanbul Technical University Project Number: MGA-2017-40811. In this work, Furkan Ozcelik was supported by the Turkcell-ITU Researcher Funding Program. 
Authors acknowledge the support of ITU Center for Satellite Communications and Remote Sensing (ITU-CSCRS) by providing Pleiades satellite images for this research.

\begin{IEEEbiography}[{\includegraphics[width=1in,height=1.25in,clip,keepaspectratio]{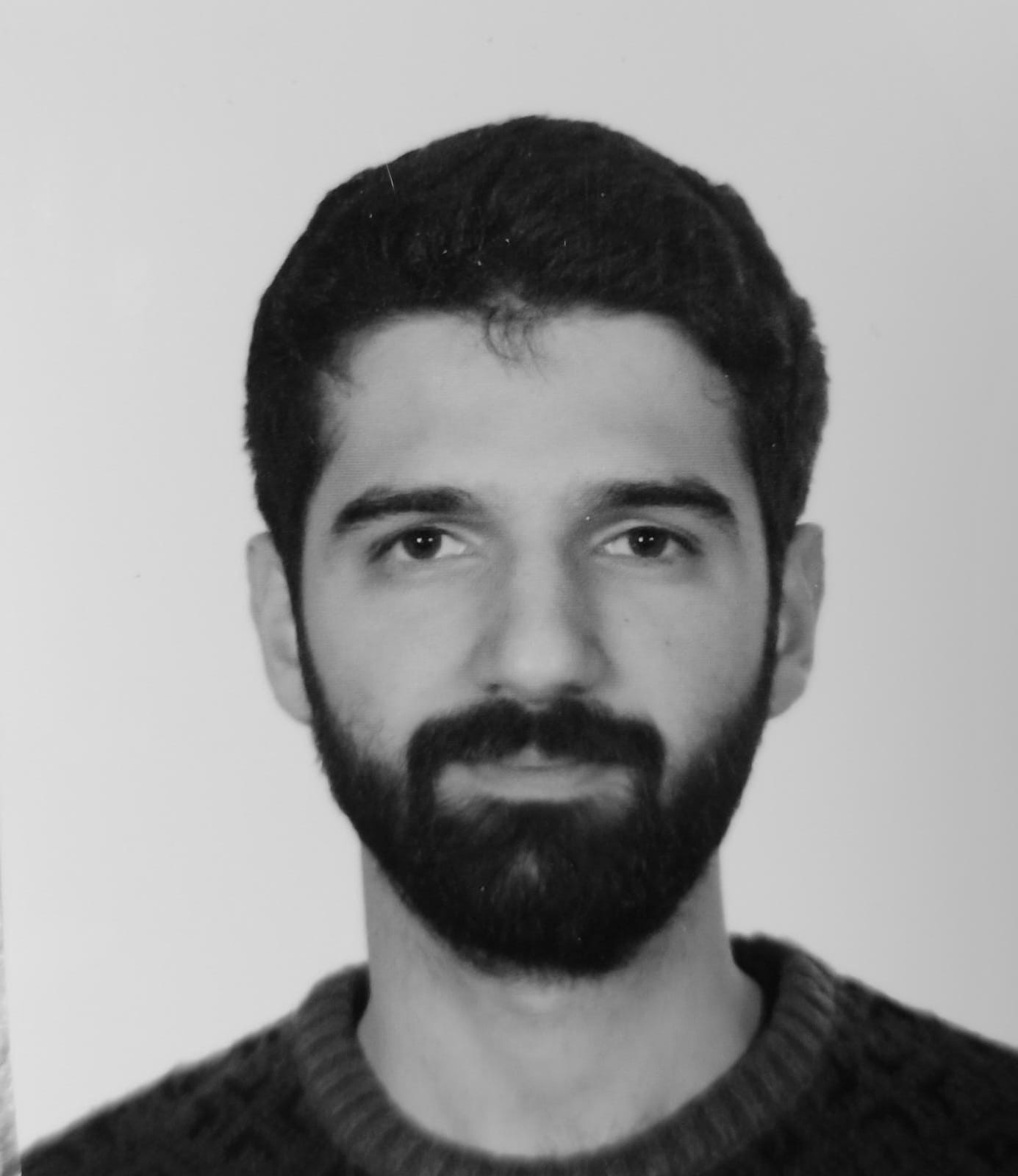}}]{Furkan Ozcelik}
 received the B.Sc. degree of Computer Engineering from Istanbul Technical University(ITU) in 2018. He was among the top 3 students in the graduating class of 2018 in ITU Faculty of Computer and Informatics. Furkan is currently pursuing his M.Sc. degree on Computer Engineering at ITU. His main research interests focus on machine learning, deep learning and generative models. 
\end{IEEEbiography}

\begin{IEEEbiography}[{\includegraphics[width=1in,height=1.25in,clip,keepaspectratio]{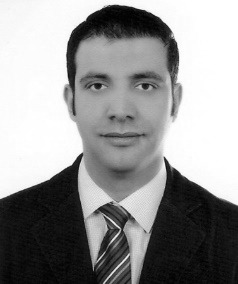}}]{Ugur Alganci}
 received the B.S., M.S. and PhD. degrees in Geomatics Engineering from the Istanbul Technical University, Istanbul, Turkey, in 2006, 2008 and 2014 respectively. 
He worked as an honorary fellow in University of Wisconsin – Madison in 2012, during his PhD researches.  He is an Assoc. Professor at Geomatics Engineering Department of Istanbul Technical University and consultor at Istanbul Technical University - Center for Satellite Communications and Remote Sensing (ITU-CSCRS).
Prof. Alganci’s research interests are remote sensing, image processing, agricultural analysis, land cover/use change. He has been reviewing for several journals in topics of remote sensing, GIS and environmental monitoring.
\end{IEEEbiography}

\begin{IEEEbiography}[{\includegraphics[width=1in,height=1.25in,clip,keepaspectratio]{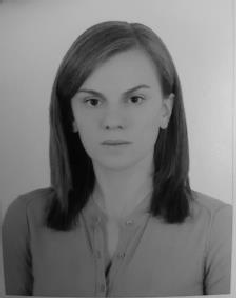}}]{Elif Sertel}
received the B.S., M.S. and PhD. degrees in Geomatics Engineering from the Istanbul Technical University, Istanbul, Turkey, in 2002, 2004 and 2008 respectively. 
She conducted her PhD dissertation work at Rutgers University, USA as a Fulbright Scholar between 2006 and 2008. She is a Professor at Geomatics Engineering Department of Istanbul Technical University and she has been holding the Director position at Istanbul Technical University - Center for Satellite Communications and Remote Sensing (ITU-CSCRS) since May 2012. 
Prof. Sertel received Outstanding Young Scientist Award from Turkish Academy of Sciences in 2017 (TÜBA-GEBİP 2017). Her main research interests are remote sensing, image processing, geostatistics, disaster management, land cover/use change, and impacts of land cover change on climate. 
\end{IEEEbiography}

\begin{IEEEbiography}[{\includegraphics[width=1in,height=1.25in,clip,keepaspectratio]{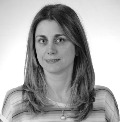}}]{Gozde Unal}
received her PhD in ECE with a minor in Mathematics from North Carolina State University, NC, USA, in 2002. After a postdoctoral fellowship at Georgia Institute of Technology, USA in 2002-2003, Dr. Unal worked as a research scientist at Siemens Corporate Research, Princeton, NJ, USA between  2003-2007. Currently, she is a full professor of Computer Engineering at Istanbul Technical University (ITU). She is the founder and member of ITU-AI Artificial Intelligence and Data Science Research and Application Center. She is the recipient of the Marie Curie Alumni Association (MCAA) Career Award 2016 of European Commission. She served as the Technical Program Co-Chair for the MICCAI Conference, 2016, Athens, and the Medical Imaging with Deep Learning Conference, 2019, London. Her main research interests are in computer vision and deep learning, particularly representation learning and generative models.
\end{IEEEbiography}







\begin{thebibliography}{10}
\providecommand{\url}[1]{#1}
\csname url@samestyle\endcsname
\providecommand{\newblock}{\relax}
\providecommand{\bibinfo}[2]{#2}
\providecommand{\BIBentrySTDinterwordspacing}{\spaceskip=0pt\relax}
\providecommand{\BIBentryALTinterwordstretchfactor}{4}
\providecommand{\BIBentryALTinterwordspacing}{\spaceskip=\fontdimen2\font plus
\BIBentryALTinterwordstretchfactor\fontdimen3\font minus
  \fontdimen4\font\relax}
\providecommand{\BIBforeignlanguage}[2]{{%
\expandafter\ifx\csname l@#1\endcsname\relax
\typeout{** WARNING: IEEEtran.bst: No hyphenation pattern has been}%
\typeout{** loaded for the language `#1'. Using the pattern for}%
\typeout{** the default language instead.}%
\else
\language=\csname l@#1\endcsname
\fi
#2}}
\providecommand{\BIBdecl}{\relax}
\BIBdecl

\bibitem{ComparisonSurvey}
G.~{Vivone}, L.~{Alparone}, J.~{Chanussot}, M.~{Dalla Mura}, A.~{Garzelli},
  G.~A. {Licciardi}, R.~{Restaino}, and L.~{Wald}, ``A critical comparison
  among pansharpening algorithms,'' \emph{IEEE Transactions on Geoscience and
  Remote Sensing}, vol.~53, no.~5, pp. 2565--2586, May 2015.

\bibitem{ComponentSubstitution}
V.~K. Shettigara, ``A generalized component substitution technique for spatial
  enhancement of multispectral images using a higher resolution data set,''
  \emph{Photogram Eng. Remote Sens}, vol.~58, no.~5, pp. 561--567, 1992.

\bibitem{MultiResolutionAnalysis}
T.~Ranchin and L.~Wald, ``Fusion of high spatial and spectral resolution
  images: The arsis concept and its implementation,'' \emph{Photogramm Eng.
  Remote Sens}, vol.~66, no.~1, pp. 49--61, January 2000.

\bibitem{MRA1}
G.~Vivone, R.~Restaino, M.~D. Mura, G.~Licciardi, and J.~Chanussot, ``Contrast
  and error-based fusion schemes for multispectral image pansharpening,''
  \emph{IEEE Geoscience and Remote Sensing Letters}, vol.~11, no.~5, p.
  930–934, 2014.

\bibitem{MRA2}
\BIBentryALTinterwordspacing
J.~Choi, H.~Park, and D.~Seo, ``Pansharpening using guided filtering to improve
  the spatial clarity of vhr satellite imagery,'' \emph{Remote Sensing},
  vol.~11, no.~6, 2019. [Online]. Available:
  \url{https://www.mdpi.com/2072-4292/11/6/633}
\BIBentrySTDinterwordspacing

\bibitem{MRA3}
X.~Meng, H.~Shen, H.~Li, L.~Zhang, and R.~Fu, ``Review of the pansharpening
  methods for remote sensing images based on the idea of meta-analysis:
  Practical discussion and challenges,'' \emph{Information Fusion}, vol.~46, p.
  102–113, 2019.

\bibitem{MTF-GLP}
B.~Aiazzi, L.~Alparone, S.~Baronti, A.~Garzelli, and M.~Selva, ``Mtf-tailored
  multiscale fusion of high-resolution ms and pan imagery,''
  \emph{Photogrammetric Engineering \& Remote Sensing}, vol.~72, no.~5, pp.
  591--596, 2006.

\bibitem{MTF-GLP-HPM}
B.~{Aiazzi}, L.~{Alparone}, S.~{Baronti}, A.~{Garzelli}, and M.~{Selva}, ``An
  mtf-based spectral distortion minimizing model for pan-sharpening of very
  high resolution multispectral images of urban areas,'' in \emph{2003 2nd
  GRSS/ISPRS Joint Workshop on Remote Sensing and Data Fusion over Urban
  Areas}, May 2003, pp. 90--94.

\bibitem{AWLP}
X.~{Otazu}, M.~{Gonzalez-Audicana}, O.~{Fors}, and J.~{Nunez}, ``Introduction
  of sensor spectral response into image fusion methods. application to
  wavelet-based methods,'' \emph{IEEE Transactions on Geoscience and Remote
  Sensing}, vol.~43, no.~10, pp. 2376--2385, Oct 2005.

\bibitem{GAN}
\BIBentryALTinterwordspacing
I.~Goodfellow, J.~Pouget-Abadie, M.~Mirza, B.~Xu, D.~Warde-Farley, S.~Ozair,
  A.~Courville, and Y.~Bengio, ``Generative adversarial nets,'' in
  \emph{Advances in Neural Information Processing Systems 27}, Z.~Ghahramani,
  M.~Welling, C.~Cortes, N.~D. Lawrence, and K.~Q. Weinberger, Eds.\hskip 1em
  plus 0.5em minus 0.4em\relax Curran Associates, Inc., 2014, pp. 2672--2680.
  [Online]. Available:
  \url{http://papers.nips.cc/paper/5423-generative-adversarial-nets.pdf}
\BIBentrySTDinterwordspacing

\bibitem{SRGAN}
C.~{Ledig}, L.~{Theis}, F.~{Huszár}, J.~{Caballero}, A.~{Cunningham},
  A.~{Acosta}, A.~{Aitken}, A.~{Tejani}, J.~{Totz}, Z.~{Wang}, and W.~{Shi},
  ``Photo-realistic single image super-resolution using a generative
  adversarial network,'' in \emph{2017 IEEE Conference on Computer Vision and
  Pattern Recognition (CVPR)}, July 2017, pp. 105--114.

\bibitem{ColorGAN}
K.~Nazeri, E.~Ng, and M.~Ebrahimi, ``Image colorization using generative
  adversarial networks,'' in \emph{Articulated Motion and Deformable Objects},
  F.~J. Perales and J.~Kittler, Eds.\hskip 1em plus 0.5em minus 0.4em\relax
  Cham: Springer International Publishing, 2018, pp. 85--94.

\bibitem{DLRSReview}
X.~X. {Zhu}, D.~{Tuia}, L.~{Mou}, G.~{Xia}, L.~{Zhang}, F.~{Xu}, and
  F.~{Fraundorfer}, ``Deep learning in remote sensing: A comprehensive review
  and list of resources,'' \emph{IEEE Geoscience and Remote Sensing Magazine},
  vol.~5, no.~4, pp. 8--36, Dec 2017.

\bibitem{PNN}
\BIBentryALTinterwordspacing
G.~Masi, D.~Cozzolino, L.~Verdoliva, and G.~Scarpa, ``Pansharpening by
  convolutional neural networks,'' \emph{Remote Sensing}, vol.~8, no.~7, 2016.
  [Online]. Available: \url{https://www.mdpi.com/2072-4292/8/7/594}
\BIBentrySTDinterwordspacing

\bibitem{PanNet}
J.~{Yang}, X.~{Fu}, Y.~{Hu}, Y.~{Huang}, X.~{Ding}, and J.~{Paisley}, ``Pannet:
  A deep network architecture for pan-sharpening,'' in \emph{2017 IEEE
  International Conference on Computer Vision (ICCV)}, Oct 2017, pp.
  1753--1761.

\bibitem{DNNPan}
W.~{Huang}, L.~{Xiao}, Z.~{Wei}, H.~{Liu}, and S.~{Tang}, ``A new
  pan-sharpening method with deep neural networks,'' \emph{IEEE Geoscience and
  Remote Sensing Letters}, vol.~12, no.~5, pp. 1037--1041, May 2015.

\bibitem{MRADNN}
A.~{Azarang} and H.~{Ghassemian}, ``A new pansharpening method using multi
  resolution analysis framework and deep neural networks,'' in \emph{2017 3rd
  International Conference on Pattern Recognition and Image Analysis (IPRIA)},
  April 2017, pp. 1--6.

\bibitem{TFNet}
\BIBentryALTinterwordspacing
X.~Liu, Q.~Liu, and Y.~Wang, ``Remote sensing image fusion based on two-stream
  fusion network,'' \emph{Information Fusion}, vol.~55, pp. 1 -- 15, 2020.
  [Online]. Available:
  \url{http://www.sciencedirect.com/science/article/pii/S1566253517308060}
\BIBentrySTDinterwordspacing

\bibitem{PSGAN}
\BIBentryALTinterwordspacing
X.~Liu, Y.~Wang, and Q.~Liu, ``{Psgan: A Generative Adversarial Network for
  Remote Sensing Image Pan-Sharpening},'' \emph{Proceedings - International
  Conference on Image Processing, ICIP}, pp. 873--877, 2018. [Online].
  Available: \url{http://arxiv.org/abs/1805.03371}
\BIBentrySTDinterwordspacing

\bibitem{TargetAdaptiveCNN}
G.~{Scarpa}, S.~{Vitale}, and D.~{Cozzolino}, ``Target-adaptive cnn-based
  pansharpening,'' \emph{IEEE Transactions on Geoscience and Remote Sensing},
  vol.~56, no.~9, pp. 5443--5457, Sep. 2018.

\bibitem{DResidualPS}
Y.~{Wei} and Q.~{Yuan}, ``Deep residual learning for remote sensed imagery
  pansharpening,'' in \emph{2017 International Workshop on Remote Sensing with
  Intelligent Processing (RSIP)}, May 2017, pp. 1--4.

\bibitem{BoostingCNN}
Y.~{Wei}, Q.~{Yuan}, H.~{Shen}, and L.~{Zhang}, ``Boosting the accuracy of
  multispectral image pansharpening by learning a deep residual network,''
  \emph{IEEE Geoscience and Remote Sensing Letters}, vol.~14, no.~10, pp.
  1795--1799, Oct 2017.

\bibitem{WaldProtocol}
\BIBentryALTinterwordspacing
L.~Wald, T.~Ranchin, and M.~Mangolini, ``Fusion of satellite images of
  different spatial resolutions: {Assessing} the quality of resulting images,''
  \emph{Photogrammetric engineering and remote sensing}, vol.~63, no.~6, pp.
  691--699, 1997. [Online]. Available:
  \url{https://hal.archives-ouvertes.fr/hal-00365304}
\BIBentrySTDinterwordspacing

\bibitem{CS-PNN}
\BIBentryALTinterwordspacing
S.~Vitale and G.~Scarpa, ``A detail-preserving cross-scale learning strategy
  for cnn-based pansharpening,'' \emph{Remote Sensing}, vol.~12, no.~3, 2020.
  [Online]. Available: \url{https://www.mdpi.com/2072-4292/12/3/348}
\BIBentrySTDinterwordspacing

\bibitem{RaGAN}
\BIBentryALTinterwordspacing
A.~Jolicoeur{-}Martineau, ``The relativistic discriminator: a key element
  missing from standard {GAN},'' in \emph{7th International Conference on
  Learning Representations, {ICLR} 2019, New Orleans, LA, USA, May 6-9, 2019},
  2019. [Online]. Available: \url{https://openreview.net/forum?id=S1erHoR5t7}
\BIBentrySTDinterwordspacing

\bibitem{Unet}
O.~Ronneberger, P.~Fischer, and T.~Brox, ``U-net: Convolutional networks for
  biomedical image segmentation,'' in \emph{Medical Image Computing and
  Computer-Assisted Intervention -- MICCAI 2015}, N.~Navab, J.~Hornegger, W.~M.
  Wells, and A.~F. Frangi, Eds.\hskip 1em plus 0.5em minus 0.4em\relax Cham:
  Springer International Publishing, 2015, pp. 234--241.

\bibitem{DCGAN}
\BIBentryALTinterwordspacing
A.~Radford, L.~Metz, and S.~Chintala, ``Unsupervised representation learning
  with deep convolutional generative adversarial networks,'' in \emph{4th
  International Conference on Learning Representations, {ICLR} 2016, San Juan,
  Puerto Rico, May 2-4, 2016, Conference Track Proceedings}, 2016. [Online].
  Available: \url{http://arxiv.org/abs/1511.06434}
\BIBentrySTDinterwordspacing

\bibitem{Pix2pix-patchgan}
P.~{Isola}, J.~{Zhu}, T.~{Zhou}, and A.~A. {Efros}, ``Image-to-image
  translation with conditional adversarial networks,'' in \emph{2017 IEEE
  Conference on Computer Vision and Pattern Recognition (CVPR)}, July 2017, pp.
  5967--5976.

\bibitem{PerceptualSim}
A.~Dosovitskiy and T.~Brox, ``Generating images with perceptual similarity
  metrics based on deep networks,'' in \emph{Proceedings of the 30th
  International Conference on Neural Information Processing Systems}, ser.
  NIPS’16.\hskip 1em plus 0.5em minus 0.4em\relax Red Hook, NY, USA: Curran
  Associates Inc., 2016, p. 658–666.

\bibitem{DigitalGlobe}
\BIBentryALTinterwordspacing
``Search our database of product samples.'' [Online]. Available:
  \url{https://www.digitalglobe.com/samples}
\BIBentrySTDinterwordspacing

\bibitem{OpenRemoteSensing}
\BIBentryALTinterwordspacing
``Open remote sensing.'' [Online]. Available:
  \url{http://openremotesensing.net/}
\BIBentrySTDinterwordspacing

\bibitem{BDSD}
A.~{Garzelli}, F.~{Nencini}, and L.~{Capobianco}, ``Optimal mmse pan sharpening
  of very high resolution multispectral images,'' \emph{IEEE Transactions on
  Geoscience and Remote Sensing}, vol.~46, no.~1, pp. 228--236, Jan 2008.

\bibitem{ATWT}
G.~{Vivone}, R.~{Restaino}, M.~{Dalla Mura}, G.~{Licciardi}, and
  J.~{Chanussot}, ``Contrast and error-based fusion schemes for multispectral
  image pansharpening,'' \emph{IEEE Geoscience and Remote Sensing Letters},
  vol.~11, no.~5, pp. 930--934, May 2014.

\bibitem{GSA}
B.~{Aiazzi}, S.~{Baronti}, and M.~{Selva}, ``Improving component substitution
  pansharpening through multivariate regression of ms $+$pan data,'' \emph{IEEE
  Transactions on Geoscience and Remote Sensing}, vol.~45, no.~10, pp.
  3230--3239, Oct 2007.

\bibitem{GLPREGFS}
G.~{Vivone}, R.~{Restaino}, and J.~{Chanussot}, ``Full scale regression-based
  injection coefficients for panchromatic sharpening,'' \emph{IEEE Transactions
  on Image Processing}, vol.~27, no.~7, pp. 3418--3431, July 2018.

\bibitem{NonlinearIHS}
M.~{Ghahremani} and H.~{Ghassemian}, ``Nonlinear ihs: A promising method for
  pan-sharpening,'' \emph{IEEE Geoscience and Remote Sensing Letters}, vol.~13,
  no.~11, pp. 1606--1610, Nov 2016.

\bibitem{SemiblindDeconv}
G.~{Vivone}, M.~{Simões}, M.~{Dalla Mura}, R.~{Restaino}, J.~M.
  {Bioucas-Dias}, G.~A. {Licciardi}, and J.~{Chanussot}, ``Pansharpening based
  on semiblind deconvolution,'' \emph{IEEE Transactions on Geoscience and
  Remote Sensing}, vol.~53, no.~4, pp. 1997--2010, April 2015.

\bibitem{TargetAdaptiveCNN_Site}
\BIBentryALTinterwordspacing
S.~Vitale, ``sergiovitale/pansharpening-cnn-python-version,'' Nov 2019.
  [Online]. Available:
  \url{https://github.com/sergiovitale/pansharpening-cnn-python-version}
\BIBentrySTDinterwordspacing

\bibitem{PanNet_Site}
\BIBentryALTinterwordspacing
 [Online]. Available: \url{https://xueyangfu.github.io/projects/iccv2017.html}
\BIBentrySTDinterwordspacing

\bibitem{Q}
{Zhou Wang} and A.~C. {Bovik}, ``A universal image quality index,'' \emph{IEEE
  Signal Processing Letters}, vol.~9, no.~3, pp. 81--84, March 2002.

\bibitem{SAM}
R.~Yuhas, A.~F.~H. Goetz, and J.~W. Boardman, ``{Discrimination among semi-arid
  landscape endmembers using the Spectral Angle Mapper (SAM) algorithm},''
  \emph{Summaries of the Third Annual JPL Airborne Geoscience Workshop, JPL
  Publ. 92–14, Vol. 1}, 1992.

\bibitem{ERGAS}
\BIBentryALTinterwordspacing
L.~Wald, ``{Quality of high resolution synthesised images: Is there a simple
  criterion ?}'' in \emph{{Third conference ''Fusion of Earth data: merging
  point measurements, raster maps and remotely sensed images''}}, T.~Ranchin
  and L.~Wald, Eds.\hskip 1em plus 0.5em minus 0.4em\relax Sophia Antipolis,
  France: {SEE/URISCA}, Jan. 2000, pp. 99--103. [Online]. Available:
  \url{https://hal.archives-ouvertes.fr/hal-00395027}
\BIBentrySTDinterwordspacing

\bibitem{sCC}
\BIBentryALTinterwordspacing
J.~Zhou, D.~L. Civco, and J.~A. Silander, ``A wavelet transform method to merge
  landsat tm and spot panchromatic data,'' \emph{International Journal of
  Remote Sensing}, vol.~19, no.~4, pp. 743--757, 1998. [Online]. Available:
  \url{https://doi.org/10.1080/014311698215973}
\BIBentrySTDinterwordspacing

\bibitem{Q4}
L.~{Alparone}, S.~{Baronti}, A.~{Garzelli}, and F.~{Nencini}, ``A global
  quality measurement of pan-sharpened multispectral imagery,'' \emph{IEEE
  Geoscience and Remote Sensing Letters}, vol.~1, no.~4, pp. 313--317, Oct
  2004.

\bibitem{QNR}
L.~Alparone, B.~Aiazzi, S.~Baronti, A.~Garzelli, F.~Nencini, and M.~Selva,
  ``Multispectral and panchromatic data fusion assessment without reference,''
  \emph{Photogrammetric Engineering \& Remote Sensing}, vol.~74, no.~2, pp.
  193--200, 2008.

\end{thebibliography}
\end{document}